%% file: FINAL_VERSION.tex
\documentclass[journal]{IEEEtran}
\usepackage{amsmath,amsfonts}
\usepackage{algorithmic}
\usepackage{algorithm}
\usepackage{array}
\usepackage[caption=false,font=normalsize,labelfont=sf,textfont=sf]{subfig}
\usepackage{textcomp}
\usepackage{color}
\usepackage{graphicx}
\usepackage{stfloats}
\usepackage{url}
\usepackage{verbatim}

\usepackage{booktabs}
\usepackage{cite}
\usepackage{multirow}
\usepackage{amssymb}
\usepackage{makecell}

\usepackage[capitalize]{cleveref}

\input{math_commands.tex}

\begin{document}

\title{Agent-Centric Relation Graph for Object Visual Navigation}

\author{Xiaobo Hu, Youfang Lin, Shuo Wang, Zhihao Wu and Kai Lv

\thanks{
This work was supported by the National Natural Science Foundation of China under Grant 62206013. (Corresponding author: Kai Lv)

The authors are with the Beijing Key Laboratory of Traffic Data Analysis and Mining, School of Computer and Information Technology, Beijing Jiaotong University, Beijing 100044, China (e-mail: \{xiaobohu, yflin, shuo.wang, zhwu, lvkai \}@bjtu.edu.cn).

Copyright © 2023 IEEE. Personal use of this material is permitted. However, permission to use this material for any other purposes must be obtained from the IEEE by sending an email to pubs-permissions@ieee.org.
}

}

\markboth{Journal of IEEE Transactions on Circuits and Systems for Video Technology}%
{Shell \MakeLowercase{\textit{et al.}}: A Sample Article Using IEEEtran.cls for IEEE Journals}

\IEEEpubid{0000--0000/00\$00.00~\copyright~2021 IEEE}

\maketitle

\begin{abstract}
Object visual navigation aims to steer an agent toward a target object based on visual observations. 
It is highly desirable to \textit{reasonably perceive} the environment and \textit{accurately control} the agent. 
In the navigation task, we introduce an Agent-Centric Relation Graph (ACRG) for learning the visual representation based on the relationships in the environment. 
ACRG is a highly effective structure that consists of two relationships, \textit{i.e.,} \textit{the horizontal relationship among objects} and \textit{the distance relationship between the agent and objects}. 
On the one hand, we design the Object Horizontal Relationship Graph (OHRG) that stores the relative horizontal location among objects. 
On the other hand, we propose the Agent-Target Distance Relationship Graph (ATDRG) that enables the agent to perceive the distance between the target and objects. 
For ATDRG, we utilize image depth to obtain the target distance and imply the vertical location to capture the distance relationship among objects in the vertical direction.
With the above graphs, the agent can perceive the environment and output navigation actions. 
Experimental results in the artificial environment AI2-THOR demonstrate that ACRG significantly outperforms other state-of-the-art methods in unseen testing environments.
\end{abstract}

\begin{IEEEkeywords}
Object visual navigation, Relation graph, Depth estimation, Reinforcement learning.
\end{IEEEkeywords}

\section{Introduction}
\IEEEPARstart{V}{isual} navigation aims to guide the agent to the target object based on visual observations from its first perspective. 
To solve this problem, it is critical to \textit{perceive} the environment and \textit{control} the agent. 
Thus, building object relationships in the environment and designing a navigation policy are two important issues in the navigation task.
In this paper, we mainly focus on the problem of perceiving the environment and building robust and suitable relationships for the navigation policy.

Previous works \cite{mayo2021visual,du2020learning,DBLP:conf/iclr/DuY021} explicitly or implicitly learn relation graphs that reflect the object relationships. 
Mayo \textit{et al.} \cite{mayo2021visual} utilize a novel attention mechanism for navigation to preserve objects' semantic and spatial information.
Du \textit{et al.} \cite{du2020learning} propose an object relation graph to learn concurrence relationships among object classes from different environments.
Puig \textit{et al.} \cite{DBLP:conf/iclr/PuigSLWLTF021} introduce a possible location distribution for each presupposed object.
The above works only consider the relationship between environmental objects while ignoring the relationship between the agent and the target. 
However, it may be helpful to perceive the relationship between the agent and the target. 
Moreover, their method of establishing the relationship between objects is rough, and they should specifically analyze the role of the different direction relationships (horizontal, vertical, and depth) for navigation.

\begin{figure}
  \centering
  \includegraphics[width=1\linewidth]{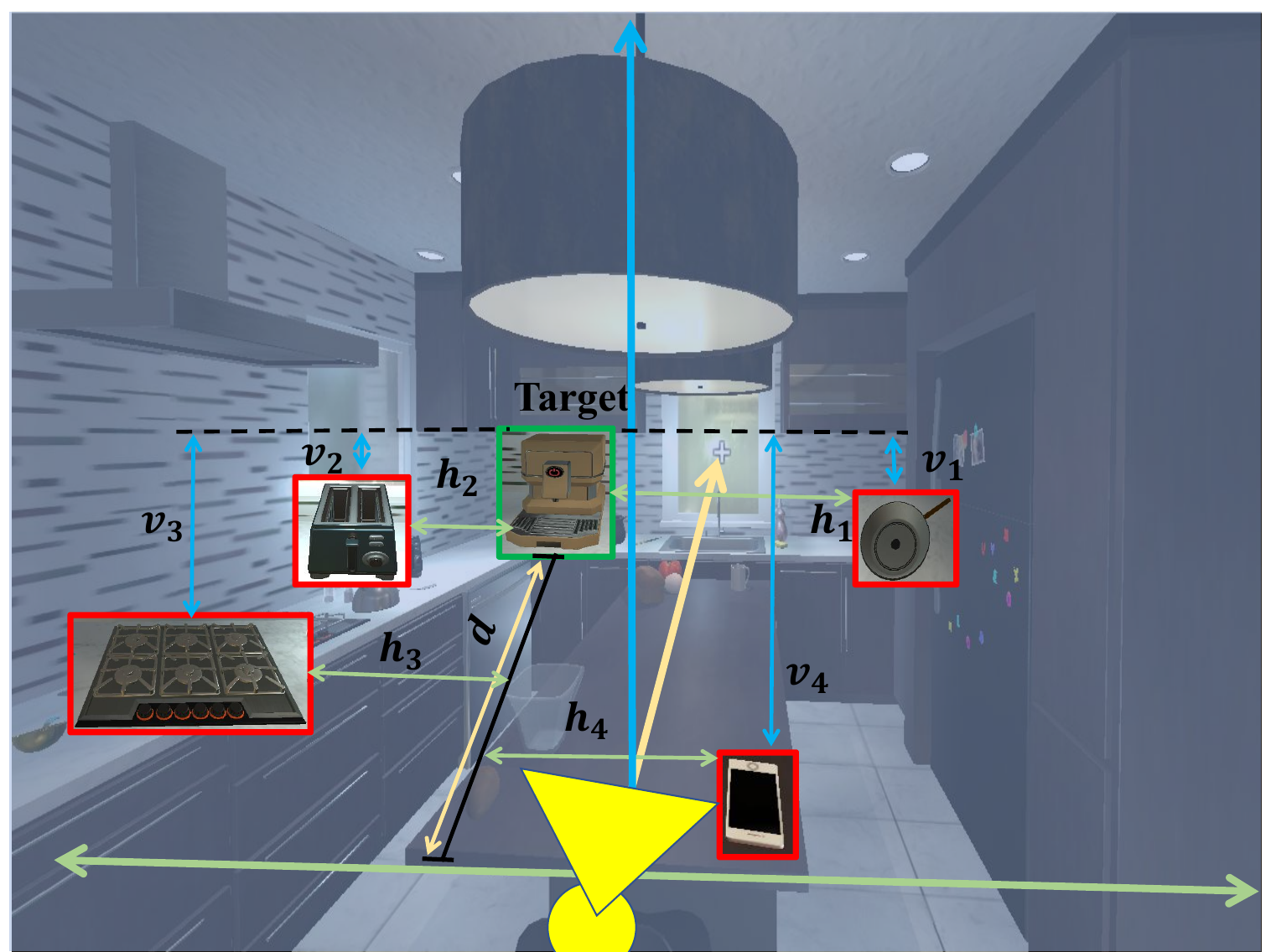}
  \caption{\textbf{Illustration of the relationships in our method. }
  The green arrow represents the horizontal direction, the blue arrow represents the vertical direction, and the yellow arrow denotes the depth distance. 
  The symbol $h_i$ represents the horizontal relationship between the target object ``CoffeeMachine'' and other objects. 
  And $d$ denotes the depth value from the target to the agent.
  The symbol $v_i$ represents the vertical relationship helping the agent to perceive the different distances among objects.
  Our model would capture the horizontal relationship among objects and the distance relationship between the agent and objects.
  }
  \label{flap_map}
\end{figure}

\IEEEpubidadjcol

In this work, we specifically analyze the impact of object relationships in different directions and propose an agent-centric model ACRG that consists of two relationships, \textit{i.e.,} \textit{the horizontal relationship among objects} and \textit{the distance relationship between the agent and objects.}
As shown in \cref{flap_map}, our model can utilize the horizontal coordinates (in Green) to capture the horizontal relationship among objects.
Moreover, our method can utilize the relationship in the vertical direction (in Blue) and the target depth value in the depth direction (in Yellow) to perceive the distance relationship. 
The construction of the above relationships is similar to the human navigation process.
For example, to ``look for a television'', humans usually perceive the target location based on the objects in the environment. 
As visual navigation is performed in an environment that contains many objects, it is reasonable to determine the rough position based on the rest of the objects, even if the target is not yet visible. 
The horizontal relationship of objects can assist in this conjecture process.
In addition, when humans approach the target, they also need to perceive the distance relationship with the target object.
We argue that the vertical relationship of objects and the depth between the agent and the target help the agent to perceive the distance relationship.
Different vertical coordinate values represent different distances from the center of the field of view. 
Visual navigation is a target-driven task, and the relationships in different directions all play a crucial role in guiding the agent to approach the target. 

To build \textit{the horizontal relationship among objects} in the environment, we propose the Object Horizontal Relationship Graph (OHRG) that only stores relative horizontal locations. 
Note that previous work \cite{du2020learning} both builds horizontal and vertical relationships.
However, we argue that using only one graph to establish two relationships simultaneously is inappropriate and cannot make each relationship play the greatest role.
In visual navigation tasks, the process of finding objects typically does not involve vertical height. 
The agent can locate the target object by searching on the two-dimensional plane.

Specifically, in searching for the target object, the agent needs to focus more on the horizontal left-right relationship among the objects on the 2D plane.
We utilize OHRG to represent the horizontal relationship among objects through an online graph network.
Specifically, we embed the horizontal positions of the detected bounding boxes to graph nodes and learn the adjacency matrix adaptively.
However, the previous approaches \cite{pennington2014glove,cartillier2021semantic} build maps using offline experience and do not generalize well.
In this work, the agent with OHRG can adaptively capture the horizontal relationship of objects from each observation and utilize it to guide movement in the scene.

To build \textit{the distance relationship between the agent and objects}, we propose the Agent-Target Distance Relationship Graph (ATDRG) that enables the agent to perceive the distance to the target. 
ATDRG focuses on building the distance relationship centered on the agent.
Firstly, we utilize a pre-trained depth estimation model to obtain the depth map $D$ from visual observation.
We employ the depth map to emphasize the distance perception of the agent from the target.
Secondly, the vertical relationship provides auxiliary information that can help the agent perceive the distance relationship among objects.
Generally speaking, objects with different height pixels tend to have implicit different distance relationships. 
Taking human vision as an example, objects at different viewing positions often implicit different distances from themselves.
When viewing a room in front view, objects near the viewing edge are closer, while objects near the viewing center are farther away.
Specifically, we utilize ATDRG to represent the distance relationship between the agent and objects using a graph network for adaptive learning.
In each node of the graph, we embed the vertical position of the detected bounding boxes and the estimated depth to learn the adjacency matrix.
ATDRG can help the agent perceive the layout relationship in the vertical direction, estimate the distance among objects and emphasize the precise distance value of the target.
With the ATDRG, the target can more accurately approach the target.

\begin{figure}
  \centering
  \includegraphics[width=1\linewidth]{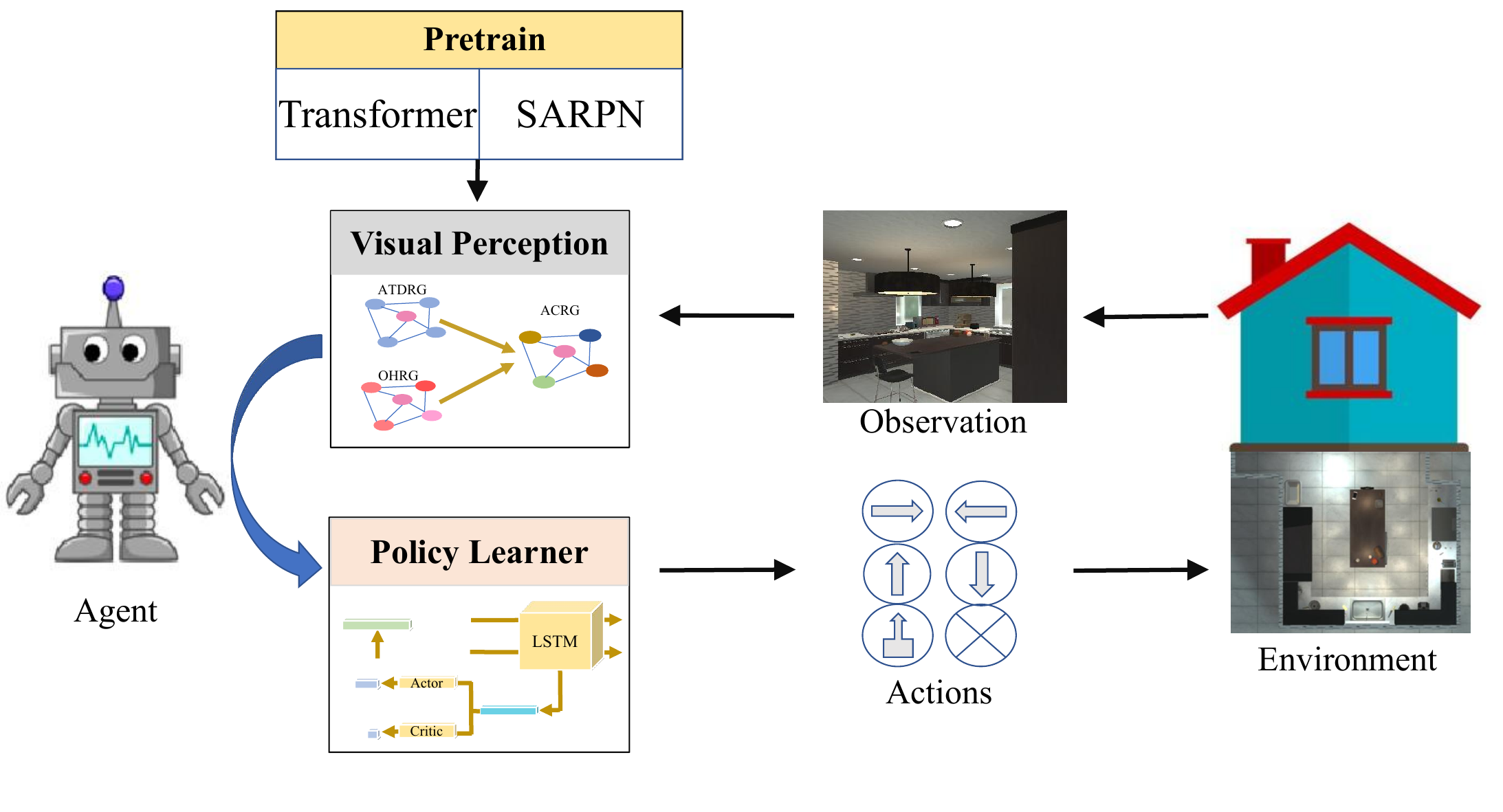}
  \caption{\textbf{Overview of our navigation framework.} Specifically, we perform feature extraction on observations through a visual perception module and then utilize a policy learner to map the features to policy actions. We also imply pre-training to accelerate the learning of the visual perception module.}
  \label{OverFlow}
\end{figure}

The overall model architecture is shown in \cref{OverFlow}.
The agent extracts features from the observation through a partially pre-trained visual perception module, then utilizes the policy learner module to map the specific policy actions.
In the visual perception module, we propose the Agent-Centric Relation Graph (ACRG) that includes OHRG and ATDRG to establish relational features from observations.
We utilize a transformer \cite{vaswani2017attention} to establish associations between graph representations and image regions to obtain the final visual representation.
In the policy learner module, we implement Long Short-Term Memory (LSTM) \cite{hochreiter1997long} to integrate previous states and utilize a standard Asynchronous Advantage Actor-Critic (A3C) architecture \cite{DBLP:conf/iclr/BabaeizadehFTCK17} to learn navigation policy. 
The reward signal is utilized for training the policy learner and visual perception simultaneously.
In addition, to facilitate the training convergence of the visual perception module, we pre-train the transformer and SARPN \cite{DBLP:conf/ijcai/ChenCZ19} model. 
The transformer is utilized to fuse features, and SARPN is implemented to generate depth value in ATDRG.
The experimental results in AI2-THOR \cite{kolve2017ai2} demonstrate the superior performance of our method. In RoboTHOR \cite{deitke2020robothor}, the results demonstrate the validity of our method in multiple environments.

In summary, this paper makes the following main points.
\begin{itemize}
\item 
We establish a horizontal object relationship in the environment and propose the Object Horizontal Relationship Graph (OHRG).
The horizontal relationship is critical in finding the target. 
\item 
We build the relationship between the agent and objects. We propose the Agent-Target Distance Relationship Graph (ATDRG) by utilizing the depth map and the vertical position of the visual observations. 
\item 
We analyze the different roles of object relationships in different directions for navigation and build an Agent-Centric Relation Graph (ACRG) by merging the above relationships.
\item 
Our method achieves new state-of-the-art accuracy on the commonly used indoor navigation simulator AI2-THOR and surpasses the existing methods by large margins.

\end{itemize}

\section{Related Work}
\subsection{Traditional Visual Navigation}
Some traditional methods in visual navigation tasks \cite{blosch2010vision,cummins2007probabilistic,dissanayake2001solution,thrun1998learning} divide navigation into three parts: mapping, positioning, and path planning. 
Thrun \textit{et al.} \cite{thrun1998learning} describe an approach that combines both paradigms, \textit{i.e.,} the grid-based and the topological.
Cummins \textit{et al.} \cite{cummins2007probabilistic} describe a probabilistic framework for navigation using only appearance data. 
In their framework, the model obtains the similarity of two observations by learning a generative appearance model.
These works require additional information and treat navigation as a purely geometric problem.
Oriolo \textit{et al.} \cite{oriolo1995line} present a practical method of online building maps automatically by fusing laser and sonar data. 
Kidono \textit{et al.} \cite{kidono2002autonomous} propose a navigation strategy that requires user assistance to make an environment map.
Other methods \cite{dissanayake2001solution,chaplot2020neural,DBLP:conf/iclr/ChaplotG00S20,shao2021mofisslam} utilize Simultaneous Localization And Mapping (SLAM) \cite{angeli2009visual} to infer scene layout and agent locations.
However, environmental maps are not always available, and even if the map is constructed, these methods are not suitable for the previously unseen environments. 

\subsection{Visual Navigation with RL}
In recent years, many approaches \cite{ng2003autonomous,kohl2004policy,mnih2015human,peters2008reinforcement,9064828} have solved the visual navigation problem through Reinforcement Learning (RL).
Generally speaking, those methods utilize visual observations as input and directly predict moving actions. 
Mirowski \textit{et al.} \cite{DBLP:conf/iclr/MirowskiPVSBBDG17} introduce prediction and loop closure classification tasks to improve navigation performance in 3D maze environments. 
Pathak \textit{et al.} \cite{pathak2018zero} stack several LSTM modules in a policy network to enhance the temporal memory, but the training process is relatively long. 
Oh \textit{et al.} \cite{9064828} present a representation learning approach considering both state and action as inputs.
Kahn \textit{et al.} \cite{kahn2018self} propose a method to build a model of the environment by a self-supervised method. 
Meanwhile, the works \cite{DBLP:conf/iclr/ChenGG19,fang2019scene} implement an intrinsic collision reward with an additional collision detector module to avoid collisions.
Other works \cite{DBLP:conf/rss/ChenVSXSVS19,DBLP:conf/iclr/SavinovDK18} introduce more information from the environment to improve navigation performance.
For example, Chen \textit{et al.} \cite{DBLP:conf/rss/ChenVSXSVS19} utilize additional topological guidance of scenes for navigation. 
In \cite{deng2009imagenet,DBLP:conf/acl/HuFRKDS19,majumdar2020improving,9265290}, natural language instructions are introduced to guide the agent.
Tang \textit{et al.} \cite{tang2021auto} introduce an auto-navigator to design a specialized network for visual navigation. 
Wu \textit{et al.} \cite{wu2019bayesian} customize a Bayesian relational memory to explore the spatial layout among rooms. 
Zhang \textit{et al.} \cite{9265290} design a cross-modal grounding module to track the correspondence between textual and visual modalities.
Shen \textit{et al.} \cite{shen2019situational} utilize multiple visual representations to generate multiple actions and then fuse those actions for obtaining an effective agent action. 

\subsection{Target-oriented Visual Navigation}
The target-oriented visual navigation model aims at steering an agent to search for different kinds of objects in an environment.
Based on the current observation and a given a specific target class, Zhu \textit{et al.} \cite{zhu2017target} employ a reinforcement learning method to generate an action. 
With the semantic segmentation and detection masks, Mousavian \textit{et al.} \cite{mousavian2019visual} propose to fuse them into the policy network for navigation. 
Wortsman \textit{et al.} \cite{wortsman2019learning} exploit Glove embedding \cite{pennington2014glove} to represent target objects and simulate the reward function for navigation in unseen environments. 
Yang \textit{et al.} \cite{DBLP:conf/iclr/YangWFGM19} implement relationships among object categories for navigation by utilizing an external knowledge database.
Veli{\v{c}}kovi{\'c} \textit{et al.} \cite{velickovic2018graph} propose a graph convolutional network to exploit the relationship between the object categories. 
However, the works \cite{DBLP:conf/iclr/YangWFGM19,velickovic2018graph} require external knowledge databases, making them unpractical in unseen environments. 
In this paper, we dynamically explore the different layout relationships of objects.

Sscnav \cite{liang2021sscnav} explicitly models scene priors using a confidence-aware semantic scene completion module to complete the scene and guide the agent's navigation planning.
Trans4Map \cite{chen2023trans4map} is an end-to-end one-stage Transformer-based framework for mapping and finally forms a semantic map of the scene.
Compared with these RGB-D models, firstly, our model does not require RGB-D data but only ordinary RGB images.
We utilize a depth estimation model to estimate the depth value of objects from images.
RGB data is obviously easier than depth data to obtain in real model scenes.
In the method of using depth information, we only use the depth value of the object to capture the distance information and do not need to build a complete semantic map of the entire scene.
Secondly, the above two schemes construct a semantic map and then guide the navigation planning learning. 
However, our method is dynamic learning of the policy and inferring the target's location through the relation graph.

ORG \cite{du2020learning} and VTNet \cite{DBLP:conf/iclr/DuY021} are proposed to solve the visual navigation task and greatly improve performance. 
ORG establishes a rough object relationship graph for navigation.
However, our method specifically analyzes the influence of the relationship between objects in different directions.
Specifically, our method builds an Agent-Centric Relation Graph (ACRG), which includes the horizontal relationship among objects and the distance relationship between the target and objects.
Our graph can help agents perceive the environment more meticulously and improve navigation performance.
VTNet adopts a transformer module to fuse spatially-enhanced local descriptors and position-encoded global descriptors through the transformer, allowing the agent to learn the relationship between instances and observation regions.
Our method utilizes a well-designed ACRG to establish object relationships, \textit{i.e.,} horizontal and distance relationship.
ACRG makes the agent aware of the various direction relationships between objects and better realize the search and navigation actions.
Although we also utilize the transformer module, the kernel of our agent is ACRG.
In our method, the transformer module is only an auxiliary module to associate the object relation and the observation regions.
Compared with VTNet, we obtain a more informative feature representation than simply detecting features from visual observation.

\section{Proposed Method}

\begin{figure*}
  \centering
  \includegraphics[width=0.8\linewidth]{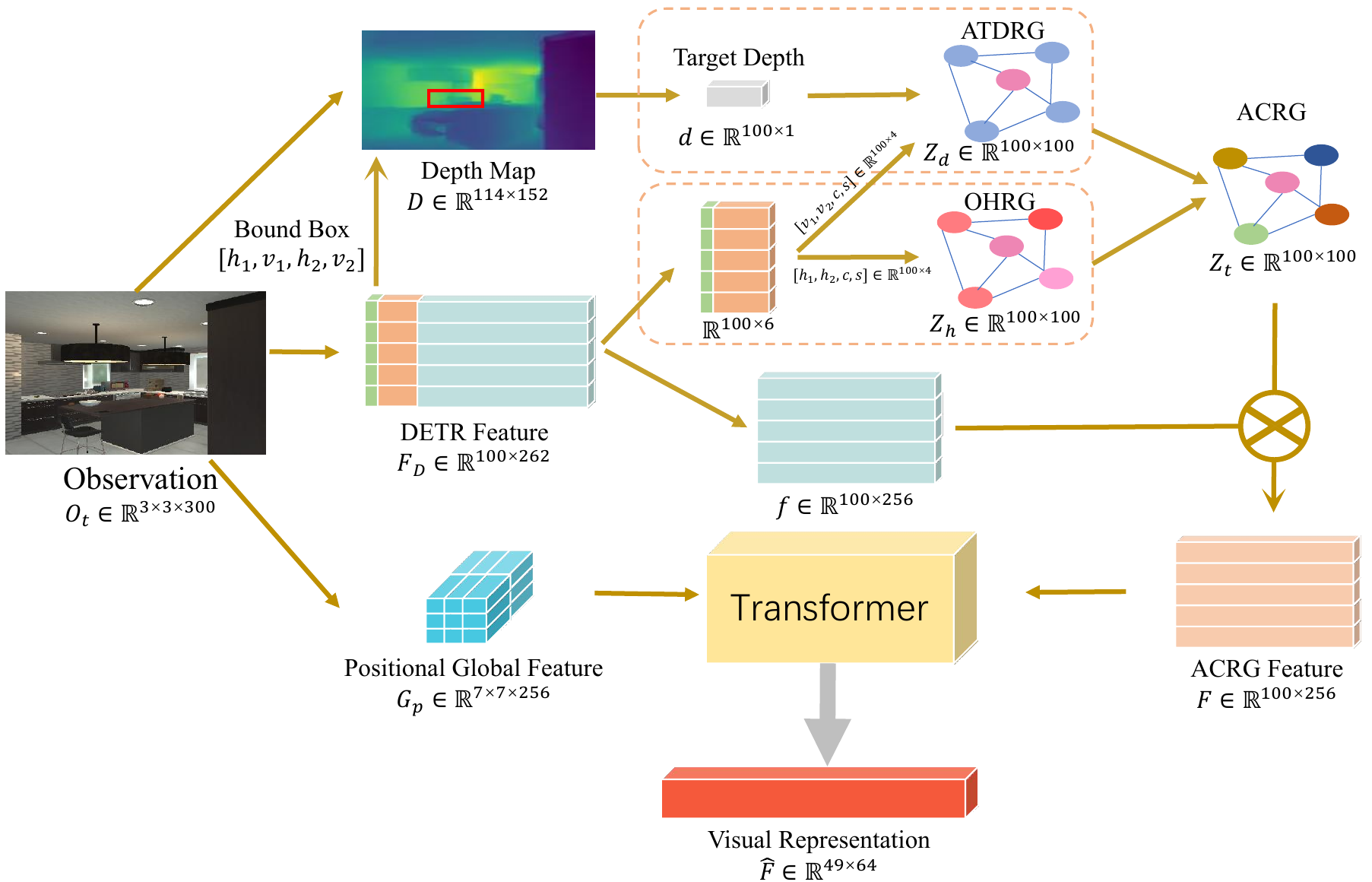}
  \caption{\textbf{Overall framework of the proposed method ACRG. }
ACRG mainly involves two different relationship graphs, \textit{i.e.}, the Object Horizontal Relationship Graph (OHRG) and Agent-Target Distance Relationship Graph (ATDRG).
On the one hand, we utilize the horizontal direction coordinates to establish OHRG.
On the other hand, we imply the target's depth value and the vertical direction coordinates to build ATDRG.
After constructing the two relational graphs and fusing the ACRG, we utilize the detection appearance feature to obtain the ACRG representation.
Then, the transformer is used to associate the ACRG representation and the global image region obtaining the final visual representation $\hat{\mF}$.
  }
  \label{Feature_Aware}
\end{figure*}

\subsection{Task Definition}
Given a target object category, such as Television, visual navigation utilizes visual observations to navigate the agent to the target object. 
The agent obtains only visual observations during navigation and does not involve additional inputs, \textit{e.g.,} scene topology map or 3D layout information.
The agent takes action based on the current observation input. 
There exist 6 types of actions in the environment, \textit{i.e.,} MoveAhead, RotateLeft, RotateRight, LookUp, LookDown, and Done.
In this work, the step size of MoveAhead is 0.25 meters, and the angles of turning left/right and looking up/down are 45$^{\circ}$
and 30$^{\circ}$, respectively. 
We define an episode as a success when (1) the agent takes the Done action, (2) the agent can observe the target in the current observation, and (3) the distance between the target and the agent is less than 1.5 meters.

At the beginning of each episode, the agent starts from a random initial state $S=(x,y,\theta _{r},\theta _{h})$ in a random room and defines a target object $T\in \left \{Television,\cdots ,Sink \right \}$.
$x$ and $y$ represent the layout coordinates of the agent, while $\theta _{r}$ and $\theta _{h}$ represent the rotation and horizontal angles of the agent camera. 
At time $t$, the agent obtains the current observation input $\tO_t$ and the previous state $\vh_{t-1}$, then takes action according to the policy $\pi (a_{t}|\tO_t,\vh_{t-1})$ to achieve the maximum value of the cumulative reward $R=\sum_{t=0}^{N_{e}}\gamma ^{t}r_{t}$. 
The state $\vh_{t-1}$ the previous environment state in reinforcement learning and has Markov properties.
The above $a_{t}$ and $r_{t}$ denote the action distribution of the agent and the reward given by the environment at time $t$, respectively. 
$\gamma$ is the discount factor, and $N_{e}$ denotes the episode length.

\subsection{Learning Visual Representation}
For the navigation task that controls an indoor agent, it is essential to learn informative visual representation. 
To obtain this representation, as shown in \cref{Feature_Aware}, we propose an Agent-Centric Relation Graph (ACRG) that consists of two parts:
\textit{Object Horizontal Relationship Graph (OHRG)} that perceives the horizontal layout relationship among objects on the planar map and \textit{Agent-Target Distance Relationship Graph (ATDRG)} that captures the distance relationship between the agent and objects.
After getting the above graph representations, we concatenate them to get the ACRG that contains the two relationships.
Then, we utilize the attention mechanism to make the agent associate the object relation and the observation regions.
Based on this idea, we utilize a transformer to process the ACRG representation and the global feature.

\subsubsection{Object Detection and Depth Estimation} 
\ 
\newline
 {\bf Object Detection.} 
To learn the relationship between the agent and the objects, we need to detect the objects from the observation. 
In this paper, we utilize DETR \cite{carion2020end} as the detector to acquire object locations.
Given an input image, DETR locates all objects of interest and converts $N$ encoded feature from the same layer to $N$ detection results through a feed-forward network.
Each detection feature $F_D$ includes a bounding box $[h_{1},v_{1};h_{2},v_{2}]$, a confidence value $c$, a semantic label $s$, and a extracted detection feature $\vf \in \R^{1 \times 256}$.
Then, $(h_{1},v_{1})$ and $(h_{2},v_{2})$ are the coordinates of the upper left corner and lower right corner, respectively.

{\bf Depth Estimation.} 
To build the relationship between the agent and the target, an intuitive method is to perceive their distance and direction.
However, the distance cannot be obtained from observation directly in visual navigation.
In this work, we utilize a pre-trained depth estimation model and introduce depth to represent the distance. 
Following SARPN \cite{DBLP:conf/ijcai/ChenCZ19}, we extract the depth map from the current visual observation.
Then, we approximate the object depth value $d$ using the mean value of the depth map $D$ within the bounding box $[h_1,v_1;h_2,v_2]$.

\subsubsection{Agent-Centric Relation Graph} 
\ 
\newline
Based on the detection and depth estimation results, we utilize the graph network $\gG=(\gV, \mA)$ to establish the agent-centric relationship, as shown in \cref{Feature_Aware}. 
$\mX \in \R^{N\times C}$ is the input of all nodes.
$\left | \mX \right | = N$ is the number of the nodes; in this paper, the N is constant and equal to 100.
$C$ denotes the dimension of the input feature. 
$\gG_h=(\gV_h, \mA_h)$ represents OHRG that captures the horizontal relationship among the objects, 
and $\gG_d=(\gV_d, \mA_d)$ denotes ATDRG that is designed to perceive the distance relationship between the agent and objects. 

{\bf Object Horizontal Relationship Graph. } 
To establish OHRG, we define a graph $\gG_h=(\gV_h, \mA_h)$, where $\gV_h$ and $\mA_h$ represent the nodes and edges, respectively. 
Note that the edge $a_{h}\in \mA_{h}$ located between two nodes expresses a horizontal direction relationship.
The input of all nodes $\gV_{h}$ is $\mX_{h}\in \R^{N\times C_h}$, including the object horizontal coordinates $[h_{1};h_{2}]$, the confidence $c$ and the semantic label $s$. 
$C_h$ denotes the dimension of the input feature.
Note the node input $\mX_{h}$ depends on the N-dimensional detection features of the observed image.
Like DETR, if the objects' number in the observation is less than N, the detector output the rest null nodes as 0.

To learn a layout relationship representation $\mZ_{h}\in \R^{N\times N}$ of all object nodes in the horizontal direction,
all the nodes $\mX_{h}$ are input to the graph convolutional network (GCN) \cite{velickovic2018graph}.
At this step, each input node is embedded by matrix $\mW_{h} \in \R^{C_h\times N}$. 
Then we embed all the nodes according to the adjacency matrix  $\mA_{h}\in \R^{N\times N}$ with each node encoding $X_{h}\cdot W_{h}$ to obtain the layout relationship $Z_{h}$.
OHRG representation is expressed as:
\begin{equation}
\mZ_{h}=ReLU(\mA_{h}\cdot \mX_{h}\cdot \mW_{h}),
\end{equation}
where $\cdot$ is the matrix multiplication operation.

Specifically, the adjacency matrix  $\mA_{h}$ is learned adaptively rather than predefined.
We treat the matrix $\mA_{h}$ as an $N\times N$ fully connected layer network parameter to adaptively learn, like $\mW_{h}$.
We feed the node input $\mX_{h}$ to the fully connected layer to achieve the effect of matrix multiplication between $\mX_{h}$ and adjacency matrix $\mA_{h}$.
We then feed the matrix product $\mX_{h}\mW_{h}$ to the fully connected layer with parameter $\mA_{h}$ and obtain the final graph convolution representation $\mZ_{h}$.

{\bf Agent-Target Distance Relationship Graph. }
To establish ATDRG, we define the graph as $\gG_{d}=(\gV_{d}, \mA_{d})$, where $\gV_{d}$ represents the nodes, and $\mA_{d}$ denotes the edges. 
The edge denotes the distance relationship between all the nodes.
We argue that the coordinates $[v_{1};v_{2}]$ in the vertical direction is conducive to learning the distance relationship among objects, and the depth value $d$ in the depth direction can emphasize the distance of the target.
Different pixel heights in the observation image represent different distances from the agent, so we introduce the vertical coordinates $[v_{1};v_{2}]$ to model the distance relationship among objects.
To perceive the depth distance between the agent and the target, we employ a depth estimation module to approximate the depth value $d$ of the target.
Note that only the target node has a real estimated depth value $d$, and the depth values of other nodes are set to $0$.
This setting can emphasize the distance judgment of the agent to the target and prevent the interference of other task-independent objects.
The node input $\mX_{d}\in \R^{N\times C_d}$ contains four parts: the object vertical coordinates $[v_{1};v_{2}]$, the depth value $d$, the semantic label $s$, and the label confidence $c$. 

Then, $\mX_{d}$ also input to the GCN to learn the distance relationship representation $\mZ_{d}\in \R^{N\times N}$  between the target and objects.
$\mW_{d} \in \R^{C_d\times N}$, and $\mA_{d}\in \R^{N \times N}$ are the matrix and the adjacency matrix, respectively.
ATDRG representation $\mZ_{d}$ can be written as:
\begin{equation}
  \mZ_{d}=ReLU(\mA_{d}\cdot \mX_{d}\cdot \mW_{d}),
\end{equation}

{\bf Map Attention and Transformer. }
After obtaining the graph representations $\mZ_{h}$ and $\mZ_{d}$, we utilize a simple yet effective method to fuse these two representations and form the final graph representation $\mZ_{t}$.
Specifically, we concatenate $\mZ_h$ and $\mZ_d$, then use a linear layer to encode the merged features.
\begin{equation}
  \mZ_{t}=ReLU(FC(\left | \mZ_{h},\mZ_{d}\right |)),
\end{equation}
where $\left|\cdot, \cdot \right|$ denotes concatenate operation, and $FC(\cdot)$ is a fully connected layer that compresses the stack graph representations. 
To make the agent pay more attention to the object appearance, we apply $\mZ_t$ as the attention map to the DETR appearance feature $\vf \in \R^{100 \times 256}$.
This process can be written as follows:
\begin{equation}
\mF=\mZ_{t}\cdot \vf.
\end{equation}

\begin{figure}
  \centering
  \includegraphics[width=1\linewidth]{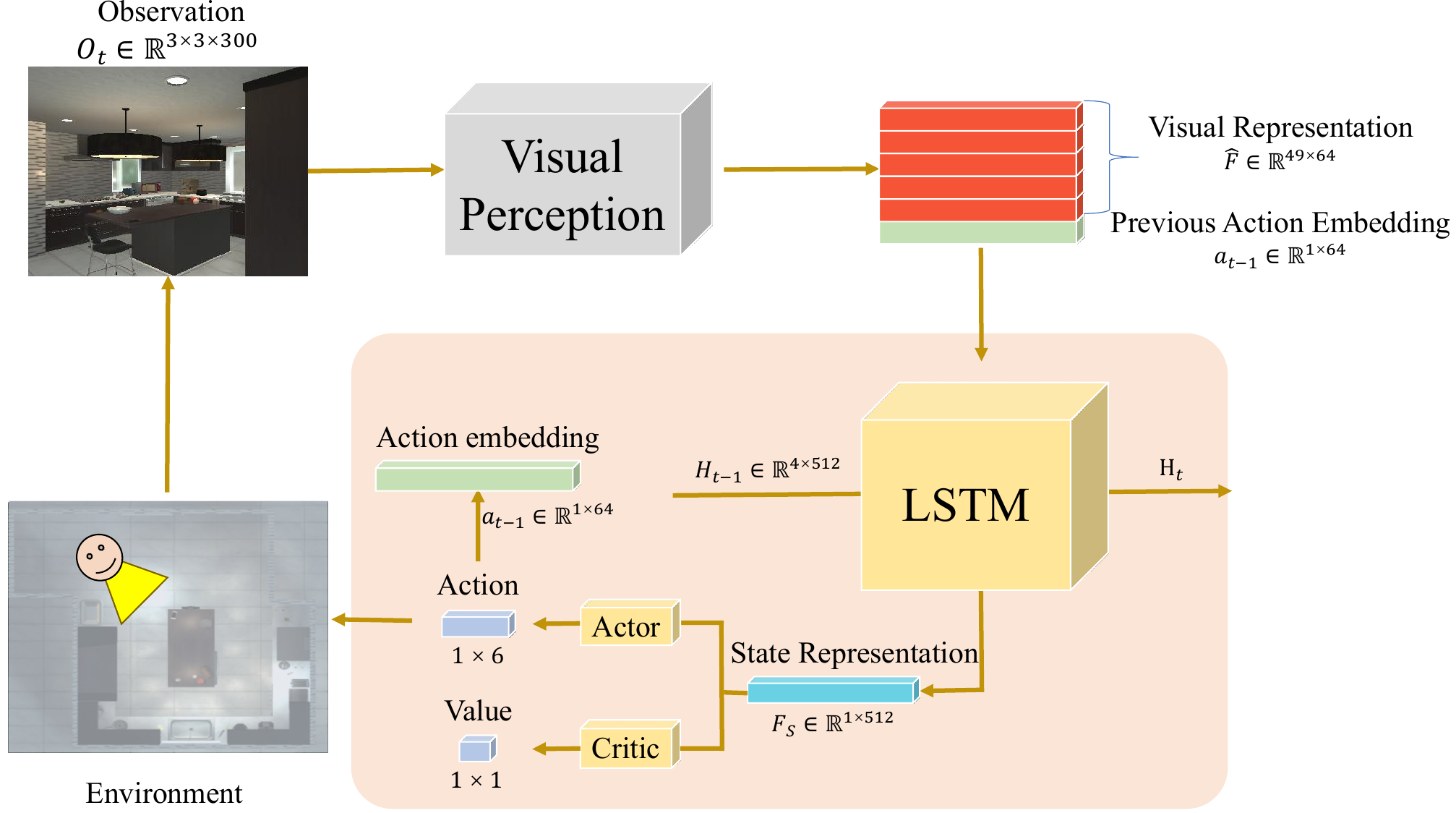}
  \caption{\textbf{Policy Learner Architecture. }
  The navigation policy learner adopts standard A3C architecture and
  utilize the LSTM to consider historical states.
  Policy learner outputs the action probability distribution and evaluate-value to control the movement of the agent.
  }
  \label{Control}
\end{figure}

Finally, we utilize a transformer to associate the graph representation $\mF$ with the image regions.
Specifically, we implement the ResNet18 \cite{he2016deep} pretrained on ImageNet \cite{deng2009imagenet} and positional embedding to generate the position-encoded global feature $\tG_{p} \in \mathbb{R}^{h\times w \times C_{g}}$ of the observation $\tO_t$.
The $h$, $w$ is the size of the feature map, and $C_g$ is the feature dimensions.
Then, $\tG_{p}$ is reshaped into a matrix-based representation $\mG_{p} \in \R^{hw \times C_{g}}$ for the convenience of attention.
We feed the $\mF$ into the transformer as keys and values by employing multi-head self-attention, and the position-encoded global feature $\mG_p$ is the query in the decoder.
The attention function of the transformer is as follows: 
\begin{equation}
Attention(\tG_{p},\mF) = SoftMax(\frac{\tG_{p}(\mF)^{'}}{\sqrt{C_{g}}})\mF. 
\end{equation}
After the processing of the transformer, we obtain the final informative visual representation $\hat{\mF}$.
We also conduct sufficient experiments in Section \ref{Ablation_Study} to verify various methods of modeling the three object relationships, \textit{i.e.,} horizontal, vertical, and depth.
The results prove that the visual representation  $\hat{\mF}\in \R^{C_{n} \times C_f}$ obtained by our proposed method is most beneficial to navigation tasks.
$C_n$ and $C_f$ are the model's number of feature channels and feature dimensions, respectively.

\textbf{Policy Learner.}
After obtaining the final visual representation $\hat{\mF}$, we utilize a basic reinforcement learning architecture to learn the navigation policy, as shown in \cref{Control}.
Considering the associations of the feature and action, we encode the previous action embedding $\va_{t-1} \in \R^{1 \times 64}$ and concatenate it onto the visual representation $\hat{\mF}$.
Then we implement LSTM to integrate the above representation and previous states $H_{t-1} \in \R^{4 \times 512}$, producing a state representation $\mF_{S} \in \R^{1 \times 512} $ for policy learning.
The process can be written as follows:
\begin{equation}
\mF_{S},\mH_{t}=LSTM(|\hat{\mF},\va_{t-1}|,\mH_{t-1})
\end{equation}
The $H_{t},H_{t-1}$ represent the hidden state of LSTM at different times, respectively.

The A3C is utilized to learn navigation policy and outputs the action distribution and values.
In each worker of A3C, the policy function is called Actor, and the value function is called Critic.
The Actor proposes a probability distribution over actions the agent can take based on the given state. 
The Critic evaluates the expected return for an agent acting according to a particular policy.
The agent selects the action with the highest probability value output by the Actor and executes it in the simulation environment.

\subsubsection{Pretraining Networks} 
\begin{figure}
  \centering
  \includegraphics[width=1\linewidth]{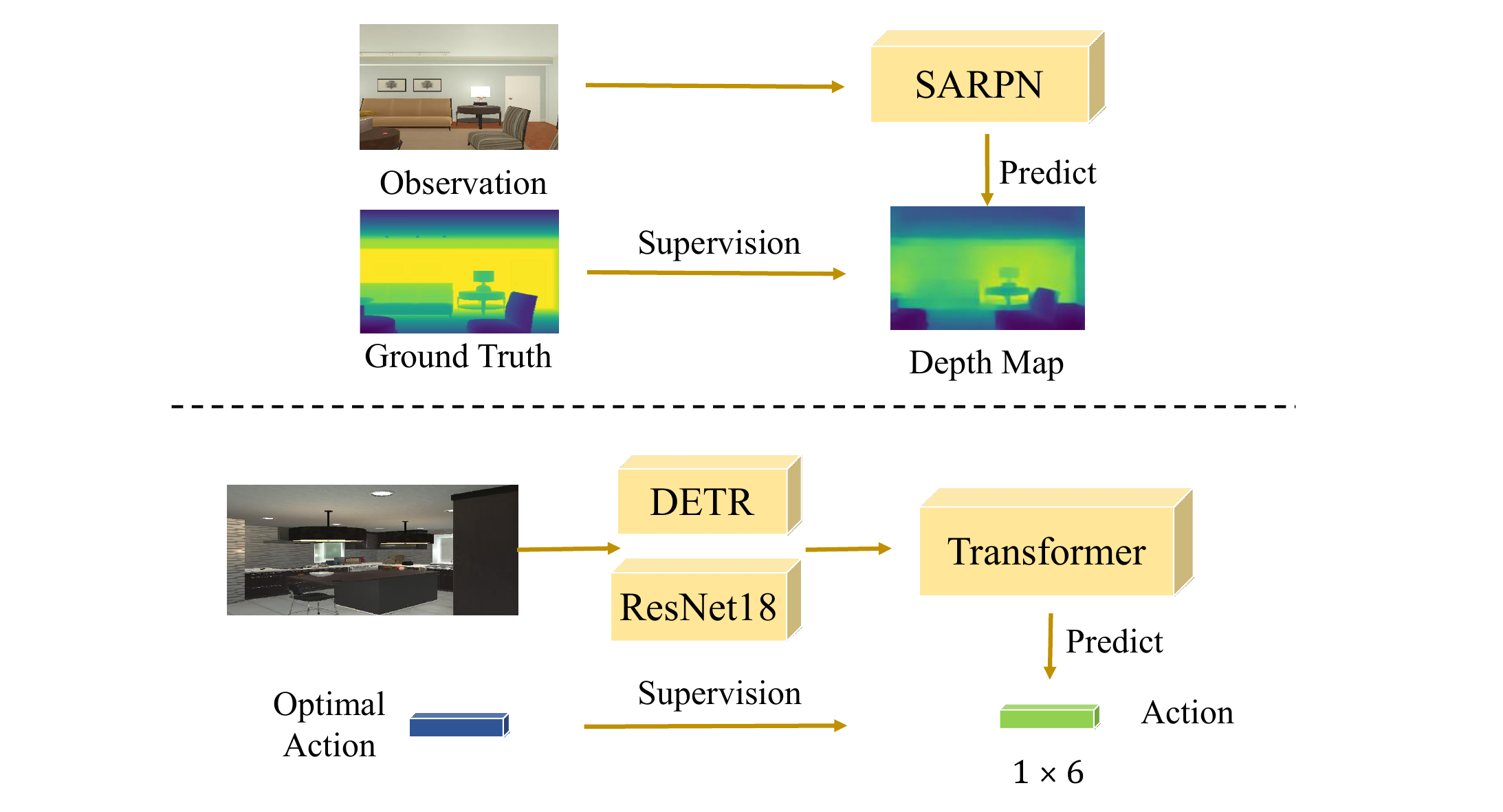}
  \caption{\textbf{Pre-training Architecture.}
  We utilize supervise learning to pre-train the two modules, \textit{i.e.,} transformer and SARPN.
  The transformer module is implied to associate the detection features with global features.
  The SARPN is a depth estimation module and implemented to generate the depth map.
}
  \label{Pretrain}
\end{figure}

\ 
\newline
The pretraining process is mainly divided into two parts, as shown in \cref{Pretrain}.
At first, we pretrain the depth estimation model SARPN \cite{DBLP:conf/ijcai/ChenCZ19} to obtain the depth map $D$ of the observation. 
We find it difficult for the depth estimation module to converge if training the depth estimation model and navigation model simultaneously.
In our method, it would provide huge noise if we utilize the above unconverged depth feature when training ATDRG, thus degrading the accuracy of the decision-making process. 
Therefore, we use offline ground truth data to pretrain the SARPN model and generate a pseudo depth dataset for offline scene data in AI2-THOR. 
With this model, ATDRG can be fed with pseudo-accurate observation depth maps.

Secondly, we pretrain the transformer in this work. 
According to \cite{DBLP:conf/iclr/DuY021}, directly feeding the decoded representation from the transformer to a navigation network would fail to learn a successful navigation policy.
The main reason is that training a deep transformer is very difficult, especially when a weak reward from reinforcement learning is given as the supervision signal. 
Therefore, the decoded feature might be uninformative and confuse the agent.
In this paper, we pretrain the transformer network to solve the difficulties of the transformer failing to converge.
Specifically, we utilize Dijkstra's Shortest Path First algorithm to generate expert actions and then skip the graph module to directly incorporate global features into the transformer for learning visual representation. 
Then, expert actions are as supervised signals, thus guiding the transformer to learn a better representation. 

\section{Experiment}
\subsection{Dataset and Evaluation}
\textbf{Dataset.}
We train and evaluate our models under AI2-THOR \cite{kolve2017ai2}, which is an artificial indoor simulation environment. 
AI2-THOR contains 4 different types of scenes, \textit{i.e.,} bedroom, kitchen, living room, and bathroom. 
Each scene contains 30 rooms, and each room has its unique furniture placement and object types.
Similar to \cite{du2020learning,DBLP:conf/iclr/DuY021}, we choose 22 categories as our target categories and ensure that there are at least four potential targets for each room category.
The agent randomly initializes a state from more than 2000 states in the environment.
We adopt the same training and evaluation protocols as  \cite{wortsman2019learning,du2020learning,DBLP:conf/iclr/DuY021,mayo2021visual} in the following experiments. 
We choose 80 out of 120 rooms as the training set and utilize the remaining 40 rooms as the test set. 
Simultaneously, we introduce the RoboTHOR \cite{deitke2020robothor} to verify the extended performance of the ACRG and SOTA models.
RoboTHOR is a more complex environment than AI2-THOR and adds a lot of compartments to make objects less visible.

\textbf{Evaluation metrics. }
We choose Success Rate SR and Success Weighted by Path Length (SPL) given in \cite{DBLP:journals/corr/abs-1807-06757} as evaluation indicators.
The SR measures the effectiveness of navigation trajectories. 
It is formulated as $\frac{1}{N_{e}}\sum_{n=0}^{N_{e}}S_{n}$, where $N_{e}$ represents the total number of episodes, $S_{n}$ is a binary indicator that defines the success of the $n_{th}$ episode.
Specifically, 1 represents a successful episode, while 0 denotes that this episode fails.
SPL represents the efficiency of navigation trajectories and is defined as  $\frac{1}{N_{e}}\sum_{n=0}^{N_{e}}S_{n}\frac{L_{opt}}{max(L_{n},L_{opt}))}$, where $L_{n}$ represents the step length of the current episode, and $L_{opt}$ represents the optimal path length of the current episode. 

\begin{table*}\large
\begin{center}
\caption{\textbf{Comparison with other navigation models.} We report the  Success Rate (SR), Success Weighted by Path Length (SPL), and their variances by repeating experiments five times are indicated in brackets.
The result demonstrates that our ACRG has optimal navigation performance.}
\label{font-table}
\resizebox{0.75\textwidth}{!}{
\begin{tabular}{p{5cm}|cp{2.3cm}|cp{2.3cm}}
\toprule
\multirow{2}{*}{Method}  & \multicolumn{2}{c}{\textbf{ALL}} \vline& \multicolumn{2}{c}{$L_{opt}\geqslant 5$}\\[1pt] \cline{2-5}
                        & \makecell[c]{\textbf{SR}} & \makecell[c]{\textbf{SPL}}   & \makecell[c]{\textbf{SR}}  & \makecell[c]{\textbf{SPL}}\\[1pt]\hline
\midrule
        Random                  &8.0(1.3)    &0.036(0.006)  &0.3(0.1)   &0.001(0.001)  \\
        WE \cite{pennington2014glove}                     &33.0(3.5)   &0.147(0.018)  &21.4(3.0)  &0.117(0.019)  \\
        SP \cite{DBLP:conf/iclr/YangWFGM19}                      &35.1(1.3)   &0.155(0.011)  &22.2(2.7)  &0.114(0.016)  \\
        SAVN \cite{wortsman2019learning}                   &40.8(1.2)   &0.161(0.005)  &28.7(1.5)  &0.139(0.005)  \\
        ORG \cite{du2020learning}                     &65.3(0.7)   &0.375(0.008)  &54.8(1.0)  &0.361(0.009)  \\
        ORG+TPN \cite{du2020learning}                &69.3(1.2)   &0.394(0.010)  &60.7(1.3)  &0.386(0.011)  \\
        ORG (DETR)\cite{du2020learning,carion2020end}        &69.3(0.4)   &0.389(0.008)  &58.0(0.8)  &0.369(0.005)  \\
        ORG+TPN (DETR)\cite{du2020learning,carion2020end}    &70.5(0.2)   &0.401(0.004)  &60.8(0.7)  &0.390(0.007)  \\
        VTNet \cite{DBLP:conf/iclr/DuY021}                   &72.2(1.0)   &0.449(0.007)  &63.4(1.1)  &0.440(0.009)  \\
        VTNet+TPN \cite{DBLP:conf/iclr/DuY021}                &73.5(1.3)   &0.440(0.009)  &63.9(1.5)  &0.440(0.011)  \\
\bottomrule 
\textbf{ACRG (Ours)}  &\textbf{77.6(1.1)}  &0.439(0.012)  &\textbf{71.0(0.5)}  &0.423(0.007) \\\hline
\textbf{ACRG+TPN (Ours)}  &\textbf{78.2(0.9)}  &\textbf{0.457(0.007)}  &\textbf{70.4(1.4)}  &\textbf{0.447(0.011)}  \\\hline
\end{tabular}}
\end{center}
\end{table*}

\begin{table*}\large
\begin{center}
\caption{\textbf{Comparison with state-of-the-art models for different optimal path lengths.}
We report SR, SPL, and their variances are indicated in brackets.
The result demonstrates that our ACRG still has the optimal performance on tasks with longer optimal paths.
}
\label{Dif_LSPL}
\resizebox{0.75\textwidth}{!}{
\begin{tabular}{p{2.5cm}|cp{1.5cm}|cp{1.5cm}|cp{1.5cm}|cp{1.5cm}}
\toprule
\multirow{2}{*}{Method}  & \multicolumn{2}{c}{$L_{opt}\geqslant 10$} \vline& \multicolumn{2}{c}{$L_{opt}\geqslant 15$}
                        \vline& \multicolumn{2}{c}{$L_{opt}\geqslant 20$}\vline& \multicolumn{2}{c}{$L_{opt}\geqslant 25$}\\[1pt] \cline{2-9}
                        & \makecell[c]{\textbf{SR}} & \makecell[c]{\textbf{SPL}}     & \makecell[c]{\textbf{SR}}  & \makecell[c]{\textbf{SPL}}
                        & \makecell[c]{\textbf{SR}} & \makecell[c]{\textbf{SPL}} & \makecell[c]{\textbf{SR}} & \makecell[c]{\textbf{SPL}}\\[1pt]\hline
\midrule

\multirow{2}{*}{ORG \cite{du2020learning}}             
&34.6   &0.229 &12.37  &0.084
&6.15   &0.040 &0.00  &0.000   \\[1pt]
&(0.41)   &(0.010) &(0.32)  &(0.010)
&(0.52)    &(0.017) &(0.000)  &(0.000)\\[1pt]\hline 

\multirow{2}{*}{VTNet \cite{DBLP:conf/iclr/DuY021} }            
&44.21   &0.277 &17.92  &0.103 
&6.18   &0.038 &2.40  &0.014\\[1pt] 
&(0.723)   &(0.019) &(0.654)  &(0.022)  
&(1.021)   &(0.012) &(1.432)  &(0.021)\\[1pt] \hline 
\bottomrule 
\multirow{2}{*}{\textbf{ACRG(Ours)}}          
&\textbf{55.8}  &\textbf{0.351}  &\textbf{32.3}  &\textbf{0.203}
&\textbf{21.2}  &\textbf{0.141}  &\textbf{9.33} &\textbf{0.067}\\[1pt]
&(0.556)  &(0.003)  &(0.917)  &(0.003) 
&(1.700)  &(0.009)  &(1.886) &(0.008)\\[1pt]\hline 
\end{tabular}}
\end{center}
\end{table*}

\subsection{Training Details}
In extracting the visual representation, we utilize a two-stage training strategy, \textit{i.e.,} \textit{pretraining components} and \textit{Learning Agent-Centric Relation Graph}. 
In pretraining the depth estimation model, we follow the work \cite{DBLP:conf/ijcai/ChenCZ19} with supervised data. 
In pretraining the transformer, we follow VTNet \cite{DBLP:conf/iclr/DuY021} and fine-tune the DETR model in the AI2-THOR scene. 
Specifically, we pretrain VTNet with a learning rate of $10^{-5}$ and adopt a learning rate decay strategy \cite{DBLP:conf/iclr/DuY021}, starting from the initial learning rate $L_{int} = 10^{-4}$.
To achieve better performance, we choose the head number of transformer equals 8, and the attention layer equals 1.
In Learning the Agent-Centric Relation Graph (ACRG), we train the navigation policy using 16 asynchronous agents with 2 million episodes. 
We implement the Adam optimizer \cite{DBLP:journals/corr/KingmaB14} to update the policy network with a learning rate of $10^{-4}$, and the maximum episode length is 50.
When the agent gets a successful episode, it receives a full reward 5. 
When the agent takes action in each episode, it receives a negative reward $-0.01$, which motivates the agent to use fewer actions to reach a success state.

 The average time for a single agent to process each image is 0.07 seconds, and it can process 13 images in one second.
We conduct model training and testing under the Unbutu 20.04.1 system. The CPU version is Hygon C86 7151 16-core Processor, and the CPU refresh rate is 120GHZ.
The graphics card we use is NVIDIA RTX A4000, the driver version is 470.82.00, and the Cuda version is 11.4.
During the training process, we use 16 agents to train a total of 2 million rounds, each agent process occupies 2077MiB of memory, and the total training time is 43 hours.

To achieve detection results from visual observations, we adopt DETR \cite{carion2020end} as the detector. 
The reason why we choose DETR rather than  Faster RCNN \cite{ren2015faster} is explained in \cite{DBLP:conf/iclr/DuY021}.
Compared with  Faster RCNN, the feature of DETR is more conducive for the agent to perform visual navigation.
The feature output by the DETR detector is embedded with position-encoded global context information, which is more suitable for feature fusion operations.
In addition, the DETR detector aligns the feature, which is output from the penultimate layer, making DETR feature more robust to scale.

In RoboTHOR, we modify the maximum episode length to 100 and the number of training episodes to 500k. 
Other training details are consistent with AI2THOR settings.

\begin{figure*}
  \centering
  \includegraphics[width=0.8\textwidth,height=0.7\textwidth]{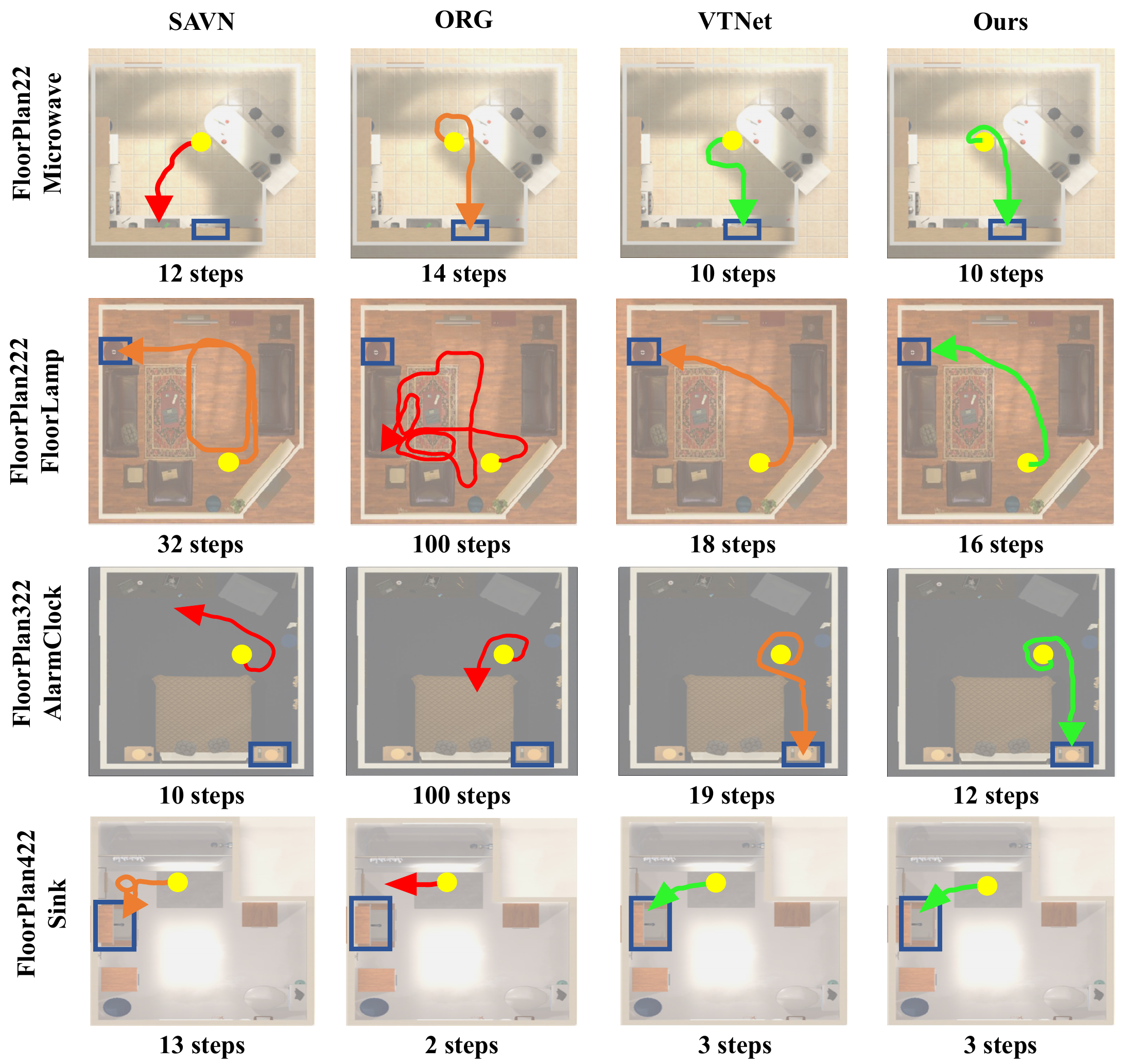}
  \caption{\textbf{Visualization results of different models in the test environment. }
  We compare the proposed method with the existing SOTA algorithms. 
  The target object is highlighted with blue bounding boxes, and the starting point is marked with a yellow dot.
  The green line represents the best case, the orange line represents the sub-optimal case, and the red represents the failed case. 
  We show four room categories and target objects, then label the room number and target category on the left side of each row.
  In most instances, SAVN cannot complete the task, ORG and VTNet takes more steps.
  Our model has the optimal navigation effectiveness and efficiency
  }
  \label{fig:casestudy}
\end{figure*}

\subsection{Comparison Methods}
{\bf Random strategy} denotes that the agent selects actions based on the average action probability. In this strategy, the agent moves randomly in the environment.

{\bf Word Embedding (WE)} \cite{pennington2014glove} uses GloVe embedding to represent the target category instead of DETR detection feature

{\bf Scene Priors (SP)} \cite{DBLP:conf/iclr/YangWFGM19} learns the category relationship among the objects from the external knowledge data FAstText \cite{DBLP:conf/eacl/GraveMJB17} and employs scene prior knowledge to navigate.

{\bf SAVN} \cite{wortsman2019learning} is a meta-reinforcement learning method that allows the agent to quickly adapt to the unseen environment.
Specifically, SAVN utilizes WEs to associate the target appearance and concept.

{\bf ORG} \cite{du2020learning} directly utilizes Faster RCNN \cite{ren2015faster} detection feature as graph nodes to establish object relationship graphs. 
ORG+TPN introduces the Tentative Policy Network (TPN) module to escape from deadlocks.  
ORG (DETR) implies DETR as the detection module, and ORG+TPN (DETR) does the same.

{\bf VTNet} \cite{DBLP:conf/iclr/DuY021} introduces a visual transformer that leverages two newly designed spatial-aware descriptors and fuses them to achieve the final visual representation.
VTNet+TPN denotes that VTNet utilizes the TPN module to improve navigation effectiveness.

\subsection{Comparison with Related Arts}
{\bf Quantitative Results. }
We demonstrate the results of the proposed method and 8 comparison methods in \cref{font-table}. 
Compared with the existing algorithms, our model increases the SR to $77.6\%$ and achieves competitive results on SPL.
Furthermore, we also introduce the TPN module to improve our navigation policy.
The result proves that ACRG+TPN achieves the optimal model effect regardless of SR or SPL.
These experimental results denote that our model achieves better visual representation, thus obviously increasing the effectiveness and efficiency of navigation. 

As shown in \cref{font-table}, our method outperforms SP and SAVN. 
SP aims to employ category relationships and uses external knowledge to encode category relationships. 
Different from SP that utilizes external knowledge data, ACRG only relies on visual observation to establish graph relationships. 
The comparison of experimental results also reflects that our graph relationship based on visual inputs has a better SR and SPL in navigation.
SAVN introduces meta-reinforcement learning, utilizes word embedding as the target indicator, and directly connects features from various modalities to generate visual representation.
Unlike SAVN, our method utilizes the agent-centric relationship graph representation, making the agent perceive the scene accurately and outperform SAVN by a large margin in terms of SR and SPL.

Although the ORG model also attempts to build relationships among objects, our method has better navigation performance on both metrics. 
ORG directly utilizes the detection results as the feature of the graph nodes and only considers perceiving the objects by building the relationship among objects in the environments. 
However, ORG ignores the distance relationship between the agent and the target and does not specifically analyze the association relationships in different directions.  
Unlike ORG, our method considers the two relationships when perceiving the environment.
The comparison results of ORG and our method clearly show that the relationships built by ACRG are more robust. 
Compared with ORG+DETR, better detection features can improve the effect of ORG.
However, using the same object detection model DETR, the navigation performance of ORG is still not as good as VTNet, and our ACRG also outperforms ORG by a large margin in terms of SR and SPL.

\begin{figure*}
  \centering
  \includegraphics[width=0.8\linewidth]{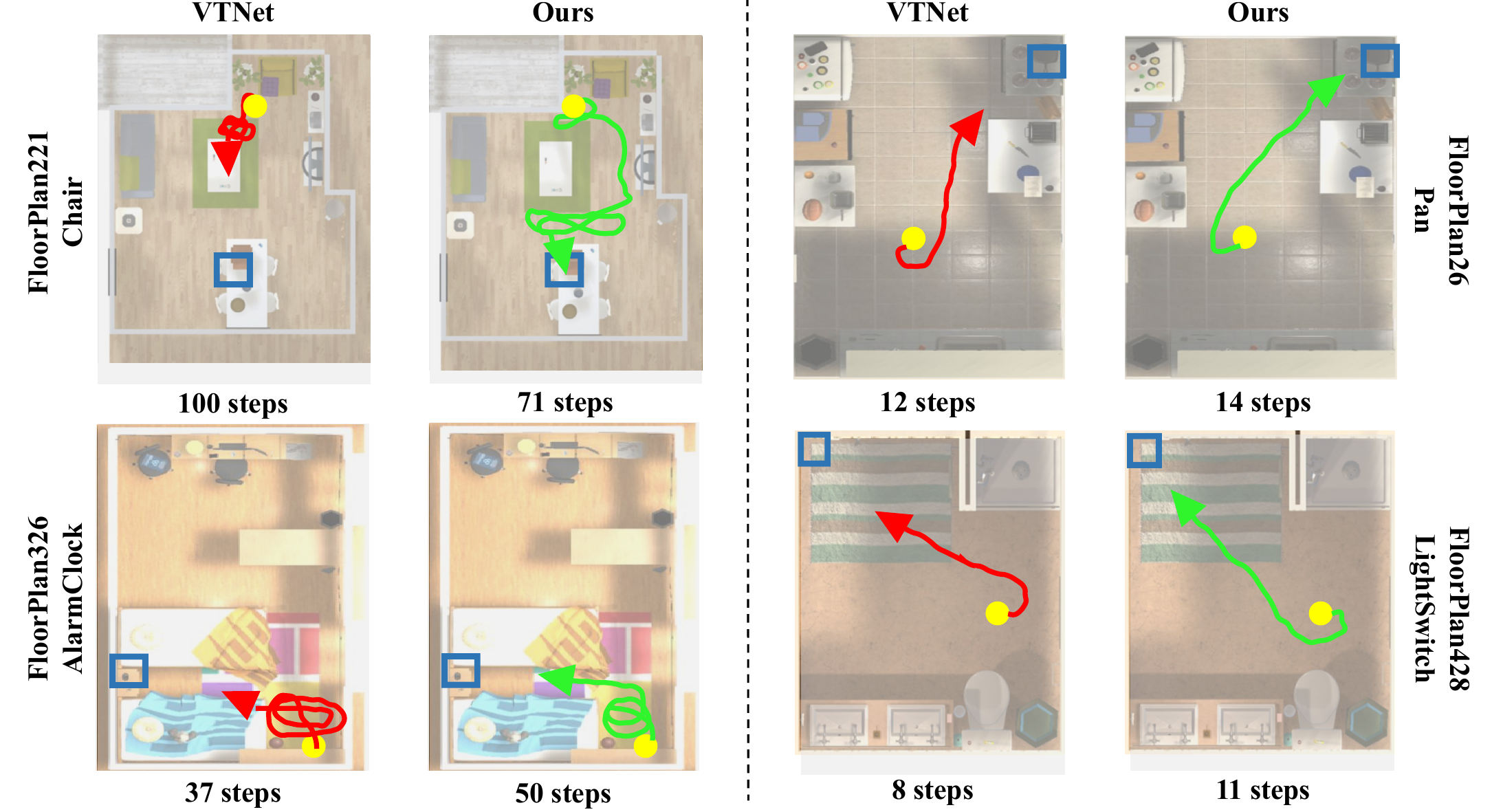}
  \caption{
  \textbf{Visualization results compared to VTNet. }
  The left side results show that our model can handle complex object layouts and eliminate obstacles compared to VTNet.
  The right shows a more accurate perception of the target distance of our model than VTNet.
  }
  \label{fig:casestudy_2}
\end{figure*}

As a SOTA algorithm, VTNet employs a transformer to decode the position-encoded global feature descriptor and the spatially enhanced local descriptor, thereby obtaining a visual representation with rich feature information. 
Our method also adopts a transformer, and the main difference between our method and VTNet is that we utilize a different visual representation derived from a graph network. 
In addition, in VTNet, only the detection and global features are used to indicate visual observations without introducing any object relationships.
These experimental results also show that building relationships between objects is important for visual navigation.

As shown in \cref{font-table}, our proposed method achieves an SPL of $0.439$, and the SPL of VTNet is $0.440$.
The results show that our method has a large SR but a competitive SPL.
To further illustrate the effectiveness of our model, we explore the SPL values under more length optimal path lengths $L_{opt}$.
As shown in \cref{Dif_LSPL}, the results show that our model can achieve better SPL on tasks with longer optimal paths.
Combining the results from the case study and RoboTHOR scene, we can conclude that our method can better perceive object layout in difficult scenes and achieve the best navigation performance.
In visual navigation, the SR reflects the effectiveness of the final result, while the SPL represents the efficiency determined by the number of steps.
We argue that the success of the plot is more important than the nuances of the amount of action.

\textbf{Qualitative Results. }
\cref{fig:casestudy} illustrates trajectories of four simple navigation tasks proceeded by four models, \textit{i.e.}, SAVN, ORG, VTNet, and our model.
We selected four different scenes, including kitchen, living room, bedroom, and bathroom, to fully illustrate the performance of each model.

The action trajectory of the SAVN model shows that this model tends to issue termination commands within a few steps of navigation, leading to mission failure or taking more redundant steps when simply navigating to the target.
This result indicates that SAVN does not extract good features from the observations and thus cannot learn a good navigation strategy.


Compared with VTNet and ORG, VTNet and our method are more efficient than ORG, and in more complex tasks.
Specifically, our method and ORG are relationship-based approaches and require building relationships. 
The visualization clearly shows that the ORG agent and our method are inclined to look around for building relationships before moving ahead. 
Our method takes fewer steps than ORG, showing that ours is more efficient in building relationships than ORG. 
VTNet relies on the real-time observations and utilizes a visual transformer to provide immediate moving actions. 
When the navigation path is short, VTNet is similar to ours.
However, when the path is longer and more complex, our agent is more efficient.


To further illustrate the validity of our model and the rationality of feature construction, our separate comparative analysis with VTNet is shown in \cref{fig:casestudy_2}.
The cases on the left are in more complex environments with obstacles in the navigation path. 
The results show that our method can successfully finish the navigation episode while VTNet cannot achieve it. 
In this case, the agent controlled by the VTNet model is stuck in a deadlock of obstacles and cannot reach the target object.
The main reason for the failure is that VTNet directly utilizes visual observation to provide moving direction and cannot perceive the layout relationship between the remaining objects from visual observations. 
In other words, when the target object appears in the observation, the agent would obtain a MovingAhead action from the VTNet model, regardless of the obstacles in the path of moving forward.
Thus, VTNet is inclined to fail in a complex environment, and our method can reach a higher SR.
Then the case on the right demonstrates the rationality of our model in building the graph.
The results show that VTNet fails the task due to incorrect distance perception of the target.
However, our model can better complete the task because it utilizes the ATDRG module to perceive the distance between the agent and objects.

\begin{table*}\large
\begin{center}
\caption{
\textbf{Comparison with different components ablation.}
We report SR, SPL, and their variances are indicated in brackets.
We compare ACRG with different ablations, \textit{i.e.,} different components, various combinations of relationships, and different methods to model the target distance.
The result demonstrates that our ACRG is the optimal solution for the navigation task.
}
\label{font-Ablation}
\resizebox{0.75\textwidth}{!}{
\begin{tabular}{p{4cm}|cp{2.3cm}|cp{2.3cm}}
\toprule
\multirow{2}{*}{Method}  & \multicolumn{2}{c}{\textbf{ALL}} \vline& \multicolumn{2}{c}{$L_{opt}\geqslant 5$}\\[1pt] \cline{2-5}
                        & \makecell[c]{\textbf{SR}} & \makecell[c]{\textbf{SPL}}     & \makecell[c]{\textbf{SR}} & \makecell[c]{\textbf{SPL}}\\[1pt]\hline
\midrule
ATDRG ($d./v.$)            &72.2(1.457)   &0.374(0.046) &61.5(1.760)  &0.309(0.023)\\[1pt]
OHRG ($h.$)            &67.8(1.375)   &0.298(0.018) &57.3(0.074)  &0.291(0.015)  \\[1pt] \hline
$h.+d.+v.$  &72.8(0.086)   &0.382(0.008)   &63.6(0.464)   &{0.375(0.004)} \\[1pt]
{$h./v.+d.$}  &75.1(0.953)   &0.421(0.011)   &65.9(1.396)   &0.407(0.005)\\ 
{$h./d./v.$}    &70.4(0.696)   &0.387(0.005)   &58.4(1.010)   &0.290(0.003)  \\[1pt]\hline
Multi-depth ACRG    &68.8(0.438)   &0.376(0.041)   &57.9(1.126)   &0.325(0.012)  \\[1pt]
Simple-depth ACRG       &71.4(0.956)   &0.395(0.010)   &60.9(1.153)   &0.296(0.121)  \\[1pt]\hline 
\bottomrule 
\textbf{ACRG ($h.+d./v.$)}  &\textbf{77.6(1.124)}  &\textbf{0.439(0.012)}  &\textbf{71.0(0.450)}  &\textbf{0.423(0.007)}  \\\hline
\end{tabular}}
\end{center}
\end{table*}

\subsection{Ablation Study \label{Ablation_Study}}
In this subsection, we compare ACRG with different component ablations and present the results in \cref{font-Ablation}.
ACRG is the proposed method, and two components are involved in our method, \textit{i.e.}, the relationship graphs OHRG and ATDRG. 
Specifically, we compare the importance of each component and the method of combining relationship graphs.

\textbf{Impact of OHRG and ATDRG. }
To verify the effectiveness of the horizontal and distance relationship modules, we compared OHRG and ATDRG modules.
The results in \cref{font-Ablation} show that only building the horizontal relationship or distance relationship would result in a lower SR and SPL than our ACRG.
Only OHRG is insufficient because the agent cannot better perceive the distance information among objects.
The representation of ATDRG only perceives the distance information but ignores the horizontal relationship among objects.
The horizontal relationship is useful for the agent to move left or right to find the target.
Without specific modeling of this relationship, the agent can only guide rough movement based on global observation.
In addition, the result also shows that ATDRG can output a better navigation performance than OHRG. 
The main reason is that ATDRG has a vertical relationship among objects and can utilize global observation to perceive the layout.
However, OHRG will not obtain the distance to the target and cannot accurately determine the conditions for success.

\textbf{Impact of combination method.}
In this section, we discuss the influence of the combination method on different layout relationships.
Specifically, ATDRG utilizes one graph to model the relationship in depth and vertical direction, and the symbol is $d./v.$.
OHRG implies one graph to model the relationship in the horizontal direction, and the symbol is $h.$.
Combining the three relationships, we considered the ACRG ($h.+d./v.$)  using one graph to model the distance relationship in depth and vertical direction and another graph to model the horizontal direction relationship.
The symbol $h./v.+d.$ means utilizing one graph to model horizontal and vertical and another to model the depth relationship.
 We also implement a combination of using three graphs to model the depth, vertical and horizontal relationship separately, and the symbol is $h.+d.+v.$.
The combination of only one graph to model the depth, vertical and horizontal, is $h./d./v.$.
We have yet to exhaustively search so that better combinations may exist.
Based on the above experimental, we propose our ACRG ($h.+d./v.$) that integrates three different direction  object relationships, \textit{i.e.}, depth, vertical and horizontal.
Our ACRG has the best performance in both SR and SPL among many variants.

The results in \cref{font-Ablation} demonstrate that only one graph establishing three relationships $h./d./v.$ or three graphs separately constructing $h. + d. + v.$ both have poor performance.
Only one graph cannot extract an effective representation from the three relationships simultaneously.
The relationships in different directions need to be established utilizing different graphs to prevent mutual interference between relationships.
Moreover, the method that employing three graphs has too many network parameters, and the captured representation is too scattered, which is not conducive to the later feature fusion.
Comparing $h./v. + d.$ and ours ACRG ($h. + d./v$), we found that using a graph to construct depth and vertical has a better performance.
Analyzing the relationship between the three directions in detail, we found that the relationship in the vertical direction can assist the agent in perceiving distance among objects.
Therefore, the vertical and depth direction relationship has commonalities, which can better guide the agent's perception of the scene layout.
In this way, our agent separately perceives the horizontal relationship for finding targets and the distance relationship for approaching targets.
There are still many possible methods for modeling the three relationships, and we will analyze this problem in the future.

\textbf{Impact of modeling depth methods. }
In this section, we analyze different methods of obtaining depth information.
In the setting of ATDRG, we only give the target node real depth value, and the other node is $0$.
We compare our ACRG to Multi-depth ACRG that gives all nodes the real depth value.
The results prove the effectiveness of only giving the depth value to the target node.
In navigation, the distance to the target is the critical factor that ultimately determines success.
We only give the real depth value to the target, which can emphasize the importance of the target distance and prevent the interference of other objects.
However, Multi-depth ACRG learns the depth relationship among objects, causing the agent to be unable to distinguish.

\begin{figure}
  \centering
  \includegraphics[width=0.93\linewidth]{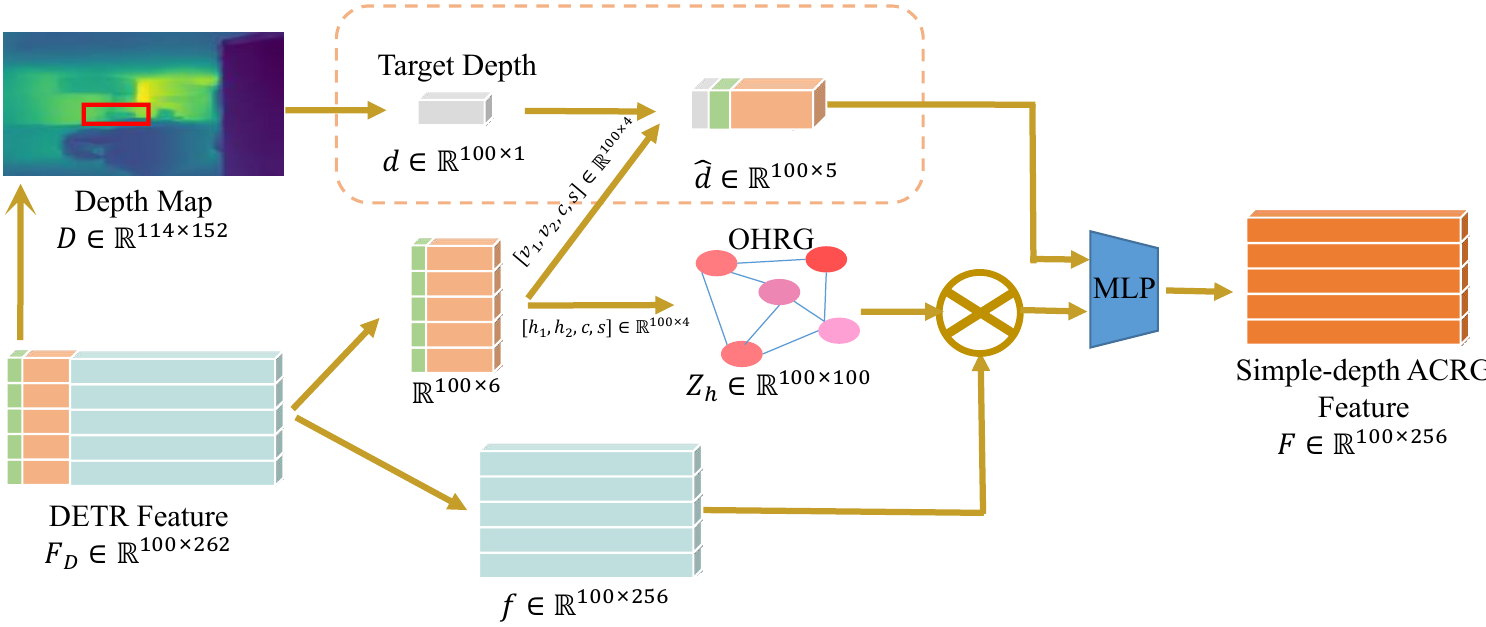}
  \caption{
  \textbf{Simple-depth ACRG Architecture.}To study the graph modeling method of distance relationship, we designed Simple-depth ACRG. 
  It doesn't utilze graph to model distance relationship and directly inputs a vector of distance information.}
  \label{Simple-depth}
\end{figure}

To investigate the effectiveness of graph modeling in perceiving distance relationships, we compared  ACRG and Simple-depth ACRG. 
As shown in Fig. 8, Simple-depth ACRG omits the use of graphs in modeling distance relationships, instead directly utilizing distance values and vertical coordinates. 
Specifically, Simple-depth ACRG takes a vector $\hat{d}$ as input, which includes the real depth value of the target $d$, vertical coordinates $[v_{1};v_{2}]$, confidence $c$, and semantic label $s$.
For OHRG representation, we apply $\mZ_h$ as the attention map to the DETR appearance feature $\vf \in \R^{1 \times 256}$.
This process can be written as $\mF=\mZ_{h}\cdot \vf.$
Finally, upon obtaining the OHRG representation $\mF$ and the depth feature $\hat{d}$, we concatenate them and transform the feature dimension to match the transformer module. 
The comparison results demonstrate the necessity of building a graph for the object distance relationship. 
Our model's ATDRG graph can better integrate vertical and depth relationships.
However, directly using a vector as input cannot fuse the two relationships and has a relatively poor effect.

\begin{table}
\begin{center}
\caption{
\textbf{Impacts of the transformer module.}
We compare variants of our method that associate the graph representation with image regions, such as siamese attention, multi-head attention (MHA), and a version without the transformer. 
We also compare the MHA and transformer with and without pre-training. 
The symbol ``w/o'' stands for ``without,'' and ``w.'' stands for ``with.'' 
The results, regarding SR, SPL, and their variances in brackets, demonstrate that our ACRG with pre-trained transformer produces the optimal visual representation and performs best.
}
\label{transformer}
\resizebox{0.5\textwidth}{!}{
\begin{tabular}{p{3.2cm}|cp{1.5cm}|cp{1.5cm}}
\toprule
\multirow{2}{*}{Method}  & \multicolumn{2}{c}{\textbf{ALL}} \vline& \multicolumn{2}{c}{$L_{opt}\geqslant 5$}\\[1pt] \cline{2-5}
                        & \makecell[c]{\textbf{SR}} & \makecell[c]{\textbf{SPL}}     & \makecell[c]{\textbf{SR}} & \makecell[c]{\textbf{SPL}}\\[1pt]\hline
\midrule
siamese attention            &60.0(0.262)   &0.272(0.004)  &47.7(1.802)  &0.256(0.006)  \\[1pt]
MHA (w/o pre-train)    &8.9(0.124)   &0.052(0.004)  &0.0  &\makecell[c]{0.0} \\[1pt] 
MHA (w. pre-train)    &62.4(0.351)   &0.301(0.002)  &49.2(1.103)  &0.284(0.003)  \\[1pt] 
transformer (w/o pre-train)    &10.2(0.112)  &0.054(0.003)  &0.0 &\makecell[c]{0.0}    \\[1pt]
w/o transformer           &75.1(0.250)   &0.426(0.005) &66.9(0.655)  &0.403(0.005) \\[1pt]
\bottomrule 
\textbf{ACRG (Ours)}  &\textbf{77.6(1.124)}  &\textbf{0.439(0.012)}  &\textbf{71.0(0.450)}  &\textbf{0.423(0.007)} \\[1pt]\hline 
\end{tabular}}
\end{center}
\end{table}

\begin{figure}
  \centering
  \includegraphics[width=0.93\linewidth]{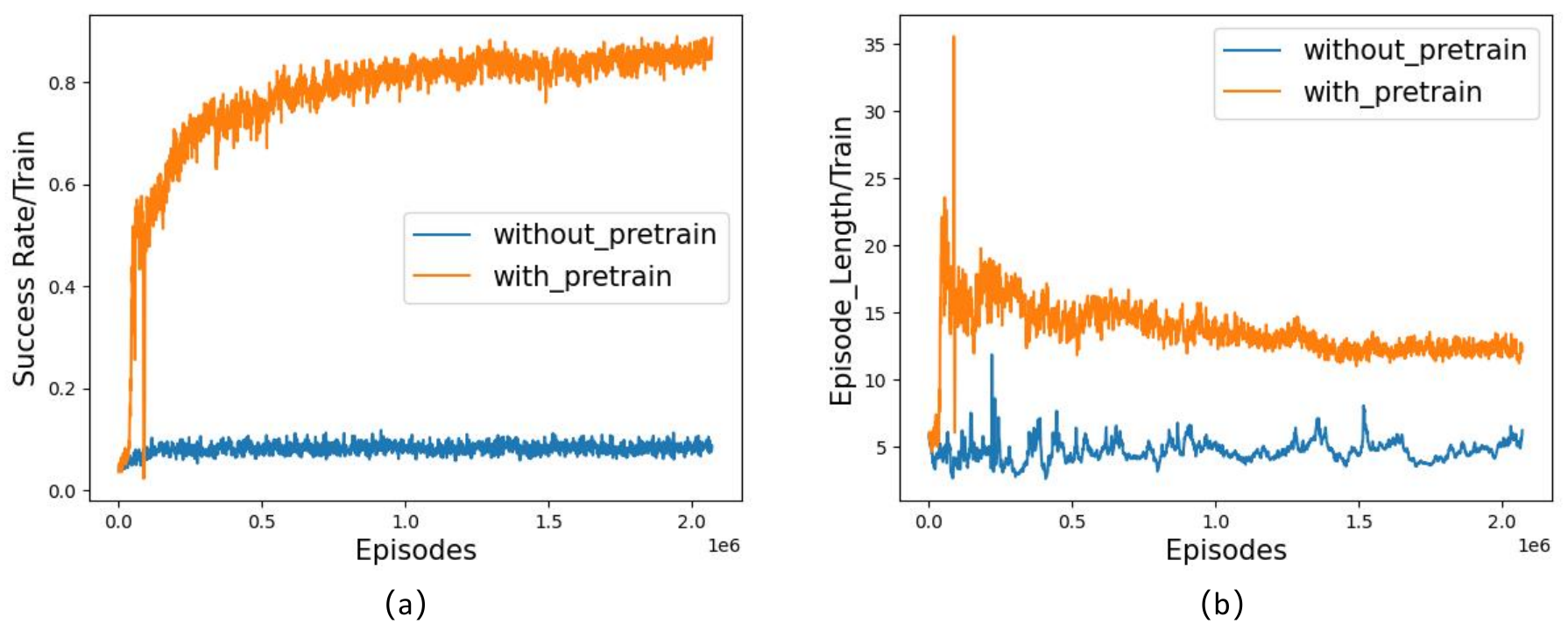}
  \caption{
  \textbf{Learning curve of pre-training phase.} We compare our model with and without the pre-training scheme. Orange and blue curves represent with and without pre-training, respectively.
  Our model will converge faster and have a higher navigation performance when using the pre-training phase.
  When without pre-training, our model will fail to converge.
  }
  \label{pretrain_cur}
\end{figure}

\subsection{Variants Study \label{Variants_Study}}

\textbf{Impact of Pre-training.}
To illustrate the effects of pre-training, we compared the curves of our model with and without the pre-training during training in Fig. \ref{pretrain_cur}.
Specifically, the success rate curve is compared in (a), and the average episode lengths are compared in (b).
The results show qualitatively that the pre-training stage improves model performance and accelerates parameter convergence.
For the success rate curve, the navigation success rate of our model with pre-training increases rapidly, while the other one cannot learn an effective navigation policy.
For the episode length curve, the model with pre-training quickly improves navigation efficiency and speeds nearly 14 steps per episode in training.
In contrast, the model without the pre-training scheme often fails to reach targets and stops around 5 steps.

\begin{table*}
\caption{
\textbf{Comparison of transformer hyperparameters.} 
We compared the results of transformers using different attention heads and network layers.
We report SR, SPL, and their variances are indicated in brackets. 
The best performance is achieved in the setting where the head number equals 8 and the attention layer equals 1.
}
\centering
\label{number}
\resizebox{0.75\textwidth}{!}{
\begin{tabular}{c|c|cc|cc}
\hline
\multirow{2}{*}{\shortstack{Head\\number}} & 
\multirow{2}{*}{\shortstack{Attention\\Layer}} & 
\multicolumn{2}{c}{ALL} \vline  & \multicolumn{2}{c}{$L_{opt}\geqslant 5$}\\[1pt] \cline{3-6}
                     &  & \makecell[c]{SR} & \makecell[c]{SPL}   &   \makecell[c]{SR}  & \makecell[c]{SPL}\\[1pt]\hline
\hline
\multirow{3}{*}{4}  
& 1  &76.7(0.616) &\textbf{0.451(0.005)} &68.6(0.250) &\textbf{0.439(0.003)}    \\[1pt]
&2  &76.5(0.205) &0.439(0.005) &67.9(0.141) &0.432(0.002)   \\[1pt]
&4  &75.8(0.500)  &0.429(0.004) &67.3(0.651) &0.424(0.002) \\[1pt]\hline
\multirow{4}{*}{8}  
&1  &\textbf{77.6(1.124)}  &0.439(0.012) &71.0(0.450) &0.423(0.007)    \\[1pt]
&2 &77.3(1.293)  &0.424(0.009) &69.6(0.367) &0.421(0.008)    \\[1pt]
&4 &77.5(1.703)  &0.421(0.018) &70.2(0.675) &0.416(0.010)   \\[1pt]
&6  &77.4(1.045) &0.405(0.015) &70.5(0.731) &0.411(0.005)    \\[1pt]
\hline 
\end{tabular}}
\end{table*}

\begin{table*}
\begin{center}
\caption{
\textbf{Comparison of computational complexity.} We show each method's floating points of operations (FLOPs) and model parameters (Params) of some important components.
The data illustrates that our model ACRG consumes only a few parameters while improving the navigation performance.
}
\label{complexity} 
\resizebox{1\textwidth}{!}{
\begin{tabular}{p{1.6cm}|cp{1.1cm}|cp{1.1cm}|cp{1.1cm}|cp{1.1cm}|cp{1.1cm}}
\toprule
\multirow{2}{*}{\textbf{Method}}  & \multicolumn{2}{c}{\textbf{ALL}} \vline& \multicolumn{2}{c}{\textbf{Transformer}}\vline& \multicolumn{2}{c}{\textbf{Graph}}\vline& \multicolumn{2}{c}{\textbf{Policy}}\vline& \multicolumn{2}{c}{\textbf{Other}}\\[1pt] \cline{2-11}
                        & \makecell[c]{\textbf{FLOPs}} & \makecell[c]{\textbf{Params}}   & \makecell[c]{\textbf{FLOPs}}  & \makecell[c]{\textbf{Params}}
                         & \makecell[c]{\textbf{FLOPs}} & \makecell[c]{\textbf{Params}}   & \makecell[c]{\textbf{FLOPs}}  & \makecell[c]{\textbf{Params}}& \makecell[c]{\textbf{FLOPs}}  & \makecell[c]{\textbf{Params}}\\[1pt]\hline
\midrule
ORG \cite{du2020learning} &13169460 &\makecell[c]{9726464}	    &0	&\makecell[c]{0}	&1642168	&\makecell[c]{74931}  &9620544	&9612423	&1906748	&\makecell[c]{39110}\\
VTNet \cite{DBLP:conf/iclr/DuY021} &103435712 &11037696  &78879744 &1057280 &0	&\makecell[c]{0} &9751616 &9743495		&14804352 &\multicolumn{1}{c}{236921} \\ 
ACRG (Ours) & 72453728	&10594549	&39464960 &\makecell[c]{528896}	&14757200	&\makecell[c]{148630}	&9751616	&9743495	&8429952	&\multicolumn{1}{c}{172928}
\\\hline
\end{tabular}}
\end{center}
\end{table*}

\textbf{Impact of the transformer module.}
We conducted experiments to study the influence of the transformer, as shown in \cref{transformer}.
We utilize different methods to associate the graph representation and the image region, \textit{i.e.}, siamese attention, multi-head attention (MHA) and without transformer.
The siamese attention and MHA are utilized to replace the transformer modules.
In the method without the transformer (w/o transformer), we directly change the dimension of the graph representation and the global feature and perform concatenation operations to obtain the features for policy learning.
In addition, we also compared the impact of pre-training on MHA and transformer \textit{i.e.,} MHA without pre-train (MHA w/o pre-train), MHA with pre-train (MHA w. pre-train),  transformer (without pre-training), and our ACRG (with pre-training transformer).
Comparing our ACRG with the ``w/o transformer'' shows that the transformer can effectively help model improve the performance for associated features.
However, the transformer only plays an auxiliary role, improving the success rate from 75\% to 77\% in formal training.
Comparing the other method shows that too many parameters of attention that have not been pre-trained will cause the model to fail to converge.
Specifically, the transformer and MHA have larger parameters. 
When pre-training is not performed, the model training will be disturbed, and they cannot learn effective navigation actions.

\textbf{Impact of transformer hyperparameters.}
We compare the effect of transformer hyperparameters on model performance in Table \ref{number}.
We vary the number of heads in multi-head self-attention and the number of layers in the transformer.
We have yet to do an exhaustive search so that better combinations may exist.
The model fails to converge to an optimal policy when the transformer layers become too deep. 
The effect of the model increases when increasing the number of attention heads.
The highest success rate is achieved when our model contains eight heads and one network layer.

\subsection{Complexity Analysis}
We show the computational complexity of the whole framework and some important components in Table \ref{complexity}, including floating points of operations (FLOPs) and model parameters (Params).
The result demonstrates that our whole parameter quantity is larger than ORG but smaller than VTNet.
Specifically, the main modification of these three models is the method to learn visual representations, so their policy learning modules have a consistent number of parameters.
Compared with ORG, our model has an additional transformer module and constructs a more refined representation of relational graphs with more parameters and calculations.
Moreover, more than half of the computational complexity comes from the transformer.
Compared with VTNet, although we all utilize the transformer module, the transformer architecture is different.
We chose the hyperparameters of 8 heads and 1 layer with fewer parameters, while VTnet chose 4 heads and 2 layers.
Overall, our model increases a small computational complexity and leads to a huge improvement in navigation performance.

\begin{figure}
  \centering
  \includegraphics[width=1\linewidth]{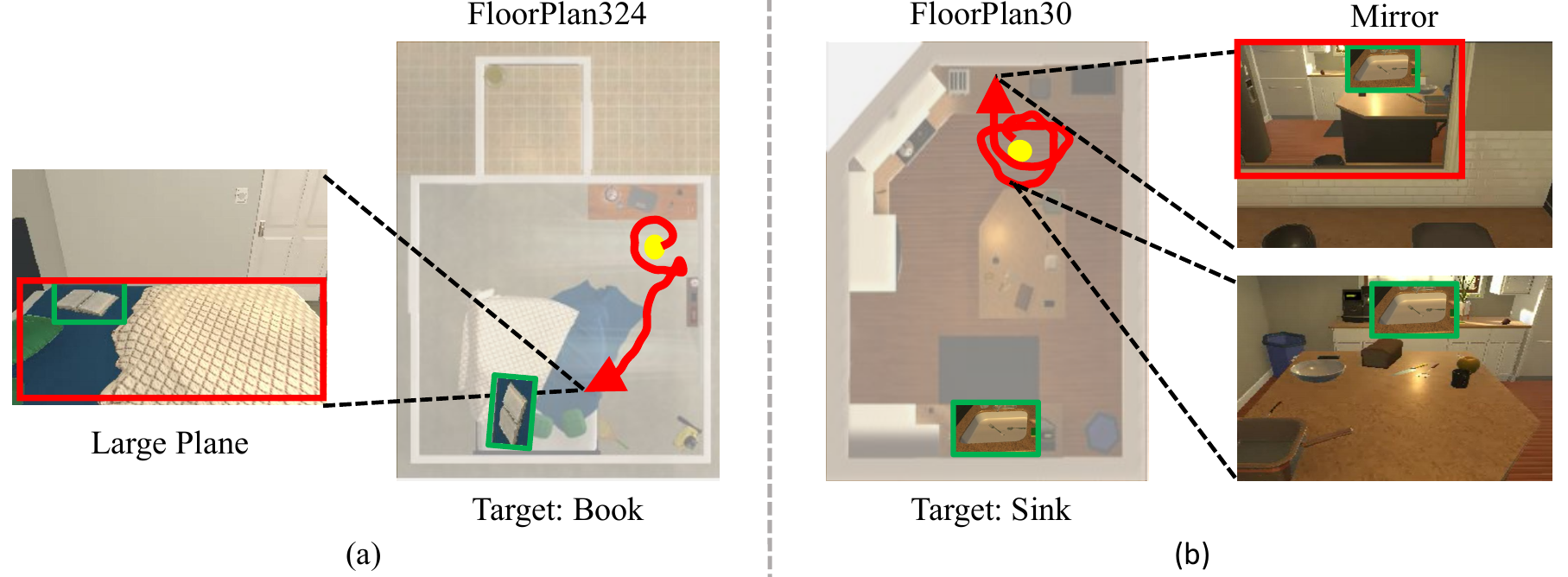}
  \caption{\textbf{Visual results of failure cases.} 
  The target object is highlighted with green bounding boxes, and the starting point is marked with a yellow dot.
  The red curve represents the failure trajectory of our agent.
  Our agents fail to reach targets due to large obstacles and mirrors interference.
  }
  \label{Falsecase}
\end{figure}

\begin{figure}
  \centering
  \includegraphics[width=1\linewidth]{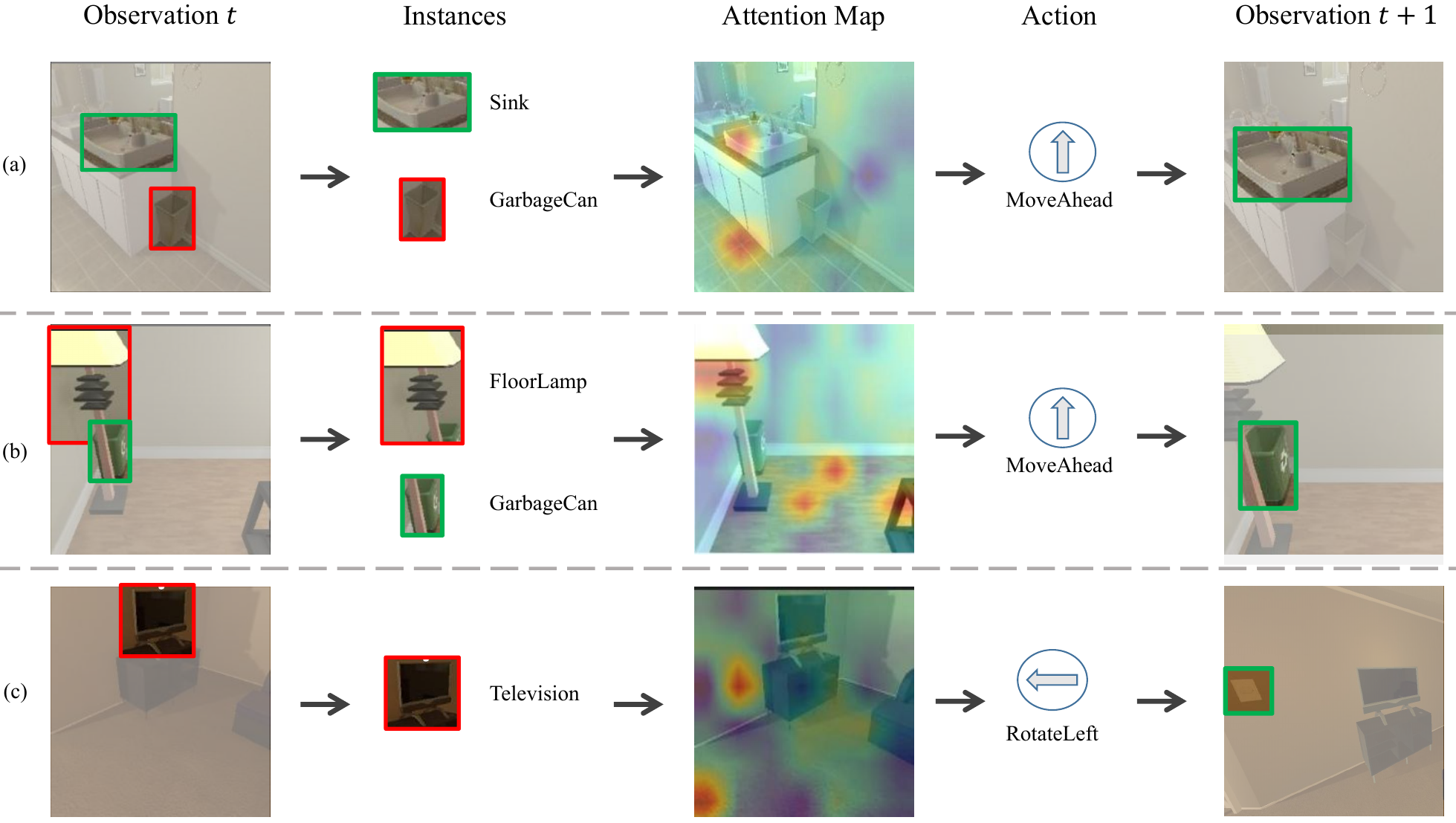}
  \caption{
  \textbf{Visualization of attention map.} 
Green bounding boxes highlight the target objects (i.e., Sink, GarbageCan, LightSwitch), and others are highlighted in red. 
The highlighted area in the figure represents the high attention value.
Our agent associates the perceived object relationship representations with the global image region. 
In (a) and (b), the agent directly observes the target object and continuously moves toward the target object.
 In (c), although the agent cannot directly observe the target, it infers the target's position from the other objects.
  }
  \label{attention_map}
\end{figure}

\subsection{Failure Cases}
As demonstrated in \cref{Falsecase}, our model fails to reach targets due to the large obstacles and specular reflections.
In \cref{Falsecase}.(a), the agent lacks the perception of the bed surface and mistakenly regards it as a possible movable area.
Our model cannot effectively perceive movable regions of indoor scenes.
Therefore, when the field of view of the agent is full of the appearance of large objects, it is impossible to effectively distinguish the object's surface from the floor texture.
For \cref{Falsecase}.(b), since the mirror reflects the room layout information, the agent finds the targets and movable paths in the mirror.
Specifically, due to the lack of perception of the mirror space, the agent only wants to reach the area displayed in mirrors based on the observation and constantly moves to mirrors.
There are often specular reflections in the home environment, and the ability to discriminate specular reflection is an important part of indoor navigation tasks based on visual observations.

\subsection{Visualization of Attention}
\cref{attention_map} demonstrates the attention maps and identifies our method can better perceive the target.
For the case in lines (a) and (b),  the object is visible, and our agent directly detects the instances of interest and then attends to the image regions.
Guided by the observation attention map, the agent selects actions to approach the targets.
In contrast, when the target object is invisible, in the case of line (c), our model can infer the approximate position of the target and attend to the regions in the image observation.
The regions corresponding to the moving direction is the potential direction where the target may exist.
The attention map illustrates that our model can obviously perceive the possible or precise location of the object in the observation image and then move toward it.

Furthermore, we compared the attention maps of VTNet and ACRG to assess their performance, as shown in Fig.12. 
In the first two cases where the agent cannot observe the target object,  ACRG focuses more on the object regions in the observed images. 
In contrast, VTNet emphasizes the movable free area in the scene. 
This difference is mainly attributed to the fact that ACRG emphasizes understanding the scene layout through the object relationship graph in different directions, while VTNet directly relies on visual input to map to optional actions. 
Moreover, in the last two cases, when the target object is visible, ACRG's prediction of the target object's location is more accurate than that of VTNet.

\begin{figure}
  \centering
  \includegraphics[width=0.9\linewidth]{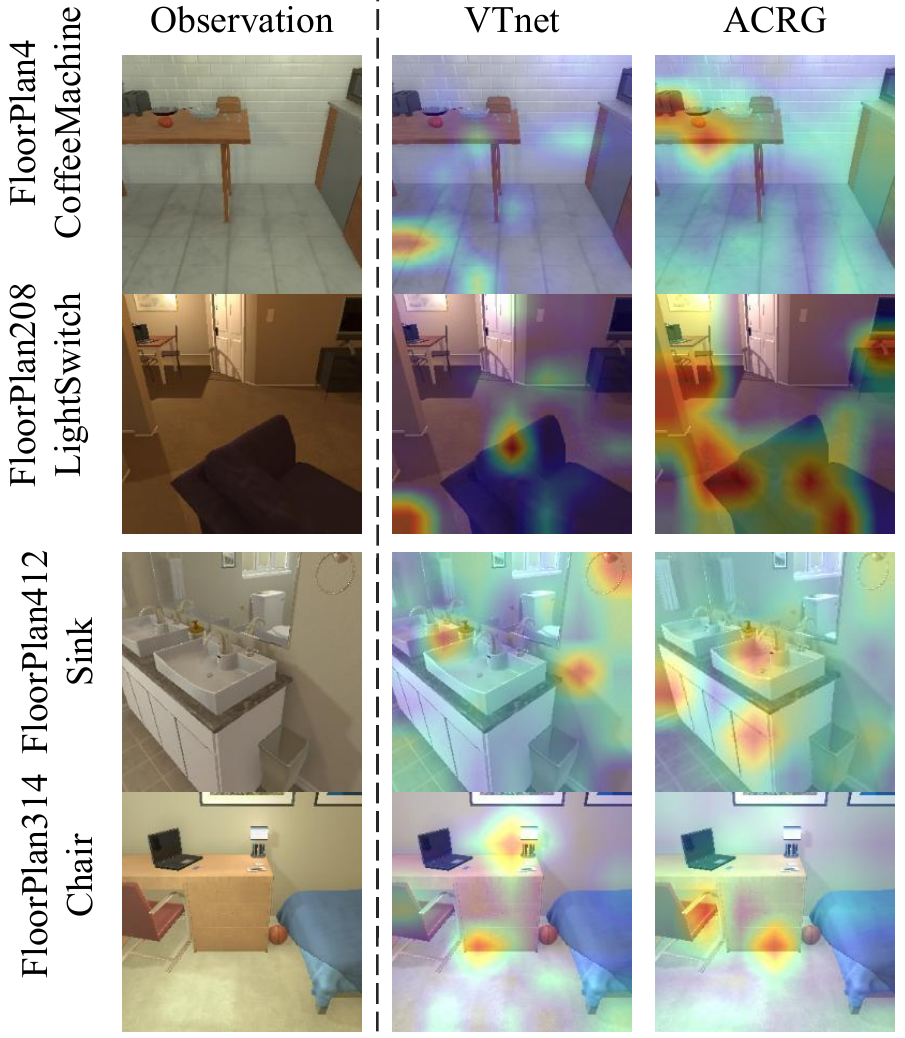}
  \caption{\textbf{Comparison of the attention maps between VTNet and ACRG.} 
  The highlighted area in the figure indicates high attention values. The results demonstrate that ACRG pays more attention to the object regions and has more precise attention to target objects than VTNet.
  }
  \label{compare_map}
\end{figure}

\subsection{Results in RoboTHOR}
\begin{figure}
  \centering
  \includegraphics[width=0.82\linewidth]{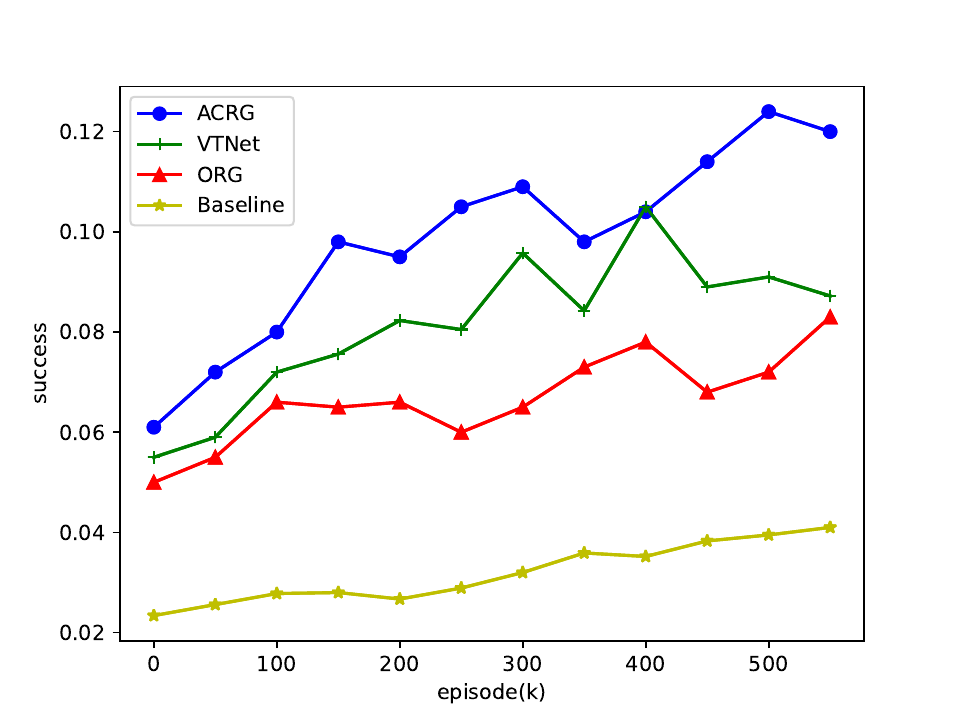}
  \caption{\textbf{Learning curve in RoboTHOR.}  
  We compare the learning curves of different training models on the RoboTHOR environment.
  The results demonstrate the effectiveness of our ACRG method.
  }
  \label{fig:expand}
\end{figure}

In this section, we discuss the performance of the navigation approaches on RoboTHOR \cite{deitke2020robothor}.
The baseline model is provided by the RoboTHOR ObjectNav 2021 Challenge.
This model utilizes RGB-D data as input and learns the policy through the DD-PPO \cite{DBLP:conf/iclr/WijmansKMLEPSB20} algorithm.
As shown in \cref{fig:expand} and \cref{expand}, we compare the learning curve in the training and the results in the testing.

As shown in \cref{fig:expand}, ACRG performs better than other models on the training set.
Specifically, the Baseline takes a lot of time to reach its expected performance.
However, ORG, VTNet, and ACRG can learn effective policies faster.
We can consider the three models efficacious in the RoboTHOR environment.
In the test, as shown in \cref{expand}, ACRG outperforms in SR and SPL.
As ORG lacks a pretraining process, its performance is slightly worse. 
Then, VTNet has a lower SR and SPL than ACRG.
Specifically, as the RoboTHOR environment contains many clapboards, the target is always invisible.
VTNet cannot perceive the target position directly through a single observation in this complex environment, 
while ACRG can utilize two relationship graphs, \textit{i.e.,} object horizontal relationship graph and agent-target depth relationship graph, to perceive the target position.

\begin{table}
\begin{center}
\caption{\textbf{ Comparison of models on RoboTHOR environment.}
We report SR, SPL, and their variances are indicated in brackets.
The result illustrates the effectiveness of our ACRG in more complex environment.
}
\label{expand}
\resizebox{0.43\textwidth}{!}{
\begin{tabular}{p{1.7cm} | cp{1cm} | cp{1cm}}
\toprule
Method  & \multicolumn{2}{c}{\textbf{SR}}\vline & \multicolumn{2}{c}{\textbf{SPL}}\\[1pt]\hline
\midrule
Baseline      &3.88  &(0.209) &0.0367 &(0.002) \\[1pt]
ORG      &9.97   &(0.676)  &0.0577  &(0.008)  \\[1pt]
VTNet            &10.4   &(0.346) &0.0655  &(0.001)  \\[1pt] \hline 
\bottomrule 
\textbf{ACRG (Ours)}  &\textbf{12.5}   &\textbf{(0.372)}&\textbf{0.0687}  &\textbf{(0.003)}\\[1pt]\hline
\end{tabular}}
\end{center}
\end{table}

\section{Conclusion}
This paper proposed a robust and effective Agent-Centric Relation Graph (ACRG) to solve the visual navigation task. 
In ACRG, we establish two relationships, \textit{i.e.,} \textit{the relationship among objects} and \textit{the relationship between the agent and the target}. 
Specifically, ACRG consists of two graphs: \textit{Object Horizontal Relationship Graph (OHRG)} and \textit{Agent-Target Distance Relationship Graph (ATDRG)}.
Benefiting from ACRG, the agent can better understand the position of objects based on visual observation.
The ACRG architecture makes the agent more effective and robust than state-of-the-art navigation models in a complex environment. 
Experiments demonstrate that our method achieves the highest performance.

{
\bibliographystyle{IEEEtran}
\bibliography{egbib}
}

\begin{IEEEbiography}[{\includegraphics[width=1in,height=1.25in,clip,keepaspectratio]{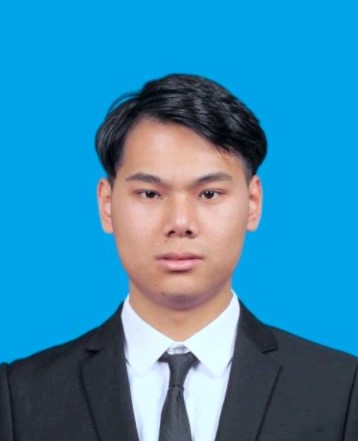}}]{Xiaobo Hu}
received the B.S. degree in computer science and technology from Beijing Jiaotong University, Beijing, China, in 2020. 

He is currently pursuing his Ph.D. candidate in the School of Computer and Information Technology at Beijing Jiaotong University, Beijing, China. 
His current research interests include computer vision, embodied intelligence, reinforcement learning.
\end{IEEEbiography}

\begin{IEEEbiography}[{\includegraphics[width=1in,height=1.25in,clip,keepaspectratio]{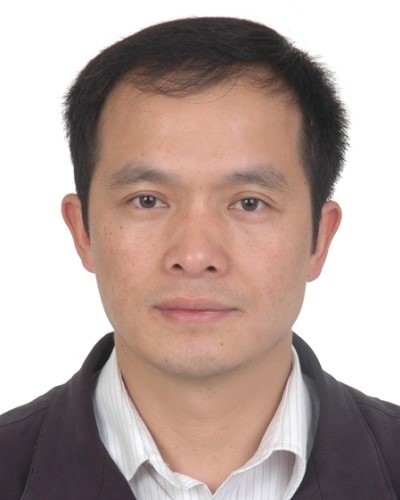}}]{Youfang Lin}
received the Ph.D. degree in signal and information processing from Beijing Jiaotong University, Beijing, China, in 2003.

He is a Professor with the School of Computer and Information Technology, Beijing Jiaotong University.
 His main fields of expertise and current research interests include big data technology, intelligent systems, complex networks, and traffic data mining.
\end{IEEEbiography}

\begin{IEEEbiography}[{\includegraphics[width=1in,height=1.25in,clip,keepaspectratio]{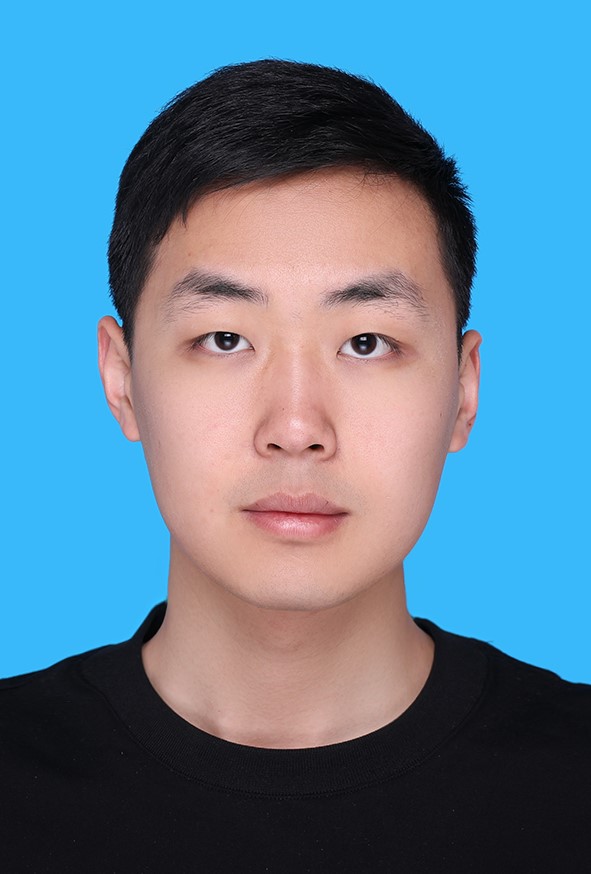}}]{Shuo Wang}
received his B.S. degree in computer science and technology from North China Electric Power University, Beijing, China, in 2017. 

He is currently pursuing his Ph.D. candidate in the School of Computer and Information Technology at Beijing Jiaotong University, Beijing, China. His research interests include computer vision, embodied intelligence, reinforcement learning, machine learning, and optimization.
\end{IEEEbiography}

\begin{IEEEbiography}[{\includegraphics[width=1in,height=1.25in,clip,keepaspectratio]{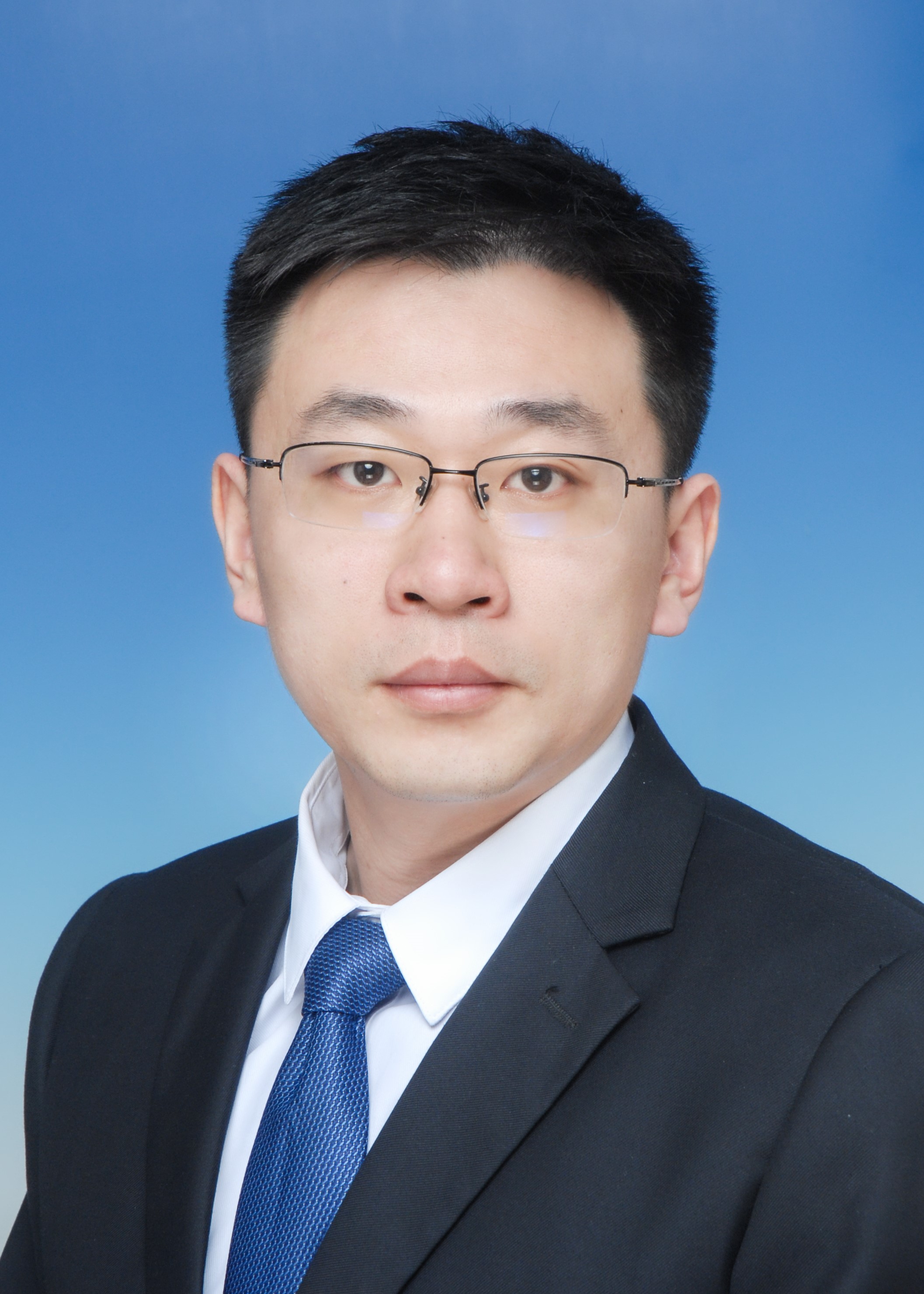}}]{Zhihao Wu}
received the Ph.D. degree in computer science and technology from Beijing Jiaotong University, Beijing, China, in 2013. 

He is currently an Associate Professor with the School of Computer and Information Technology, Beijing Jiaotong University. His current research interests focus on traffic data mining and reinforcement learning. 
\end{IEEEbiography}

\begin{IEEEbiography}[{\includegraphics[width=1in,height=1.25in,clip,keepaspectratio]{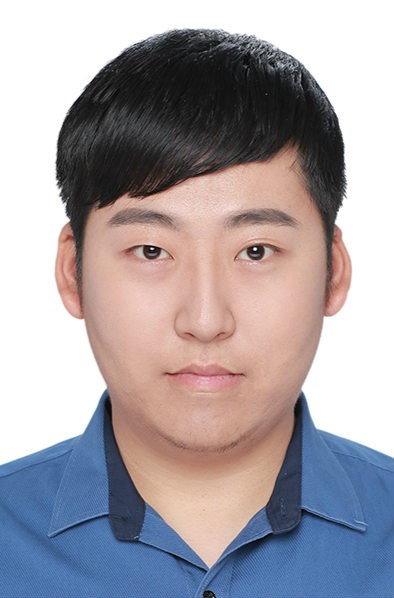}}]{Kai Lv}  received the Ph.D. degree from the School of Computer Science and Engineering, Beihang University, Beijing, China, in 2021.

 He is currently a Lecture with the School of Computer and Information Technology, Beijing Jiaotong University, Beijing, China. His main fields of expertise and current research interests include computer vision, reinforcement learning, and machine learning. He is the corresponding author of this paper.
 \end{IEEEbiography}

\end{document}

%% file: math_commands.tex
\usepackage{amsmath,amsfonts,bm}

\def\eqref#1{equation~\ref{#1}}

\def\1{\bm{1}}

\def\va{{\bm{a}}}

\def\vf{{\bm{f}}}

\def\vh{{\bm{h}}}

\def\mA{{\bm{A}}}

\def\mF{{\bm{F}}}
\def\mG{{\bm{G}}}
\def\mH{{\bm{H}}}

\def\mW{{\bm{W}}}
\def\mX{{\bm{X}}}

\def\mZ{{\bm{Z}}}

\DeclareMathAlphabet{\mathsfit}{\encodingdefault}{\sfdefault}{m}{sl}
\SetMathAlphabet{\mathsfit}{bold}{\encodingdefault}{\sfdefault}{bx}{n}
\newcommand{\tens}[1]{\bm{\mathsfit{#1}}}

\def\tG{{\tens{G}}}

\def\tO{{\tens{O}}}

\def\gG{{\mathcal{G}}}

\def\gV{{\mathcal{V}}}

\newcommand{\R}{\mathbb{R}}

